\documentclass[twoside,twocolumn,10pt]{article}

\usepackage{amssymb}
\usepackage{amsmath}

\usepackage{graphicx}
\usepackage{subcaption}
\usepackage{booktabs}
\usepackage{arydshln}
\usepackage[accsupp]{axessibility}  
\usepackage{xcolor,colortbl,multicol}

\definecolor{Gray}{gray}{0.85}
\definecolor{LightCyan}{rgb}{0.88,1,1}
\newcolumntype{a}{>{\columncolor{Gray}}c}
\newcolumntype{b}{>{\columncolor{white}}c}

\title{VLMs meet UDA: Boosting Transferability of Open Vocabulary Segmentation with Unsupervised Domain Adaptation}
\author{Roberto Alcover-Couso$^{1}$, Marcos Escudero-Viñolo$^{1}$, Juan C. SanMiguel$^{1}$ and Jesus Bescos$^{1}$ \footnote{Video Processing and Understanding Lab, Escuela Polit\'{e}nica Superior, Universidad Aut\'{o}noma de Madrid, 28049 Madrid, Spain. E-mail: roberto.alcover@estudiante.uam.es, \{marcos.escudero,juancarlos.sanmiguel, j.bescos\}@uam.es}}

\begin{document}

\maketitle
\begin{abstract}
 Segmentation models are typically constrained by the categories defined during training. To address this, researchers have explored two independent approaches: adapting Vision-Language Models (VLMs) and leveraging synthetic data. However, VLMs often struggle with granularity, failing to disentangle fine-grained concepts, while synthetic data-based methods remain limited by the scope of available datasets. 
  This paper proposes enhancing segmentation accuracy across diverse domains by integrating Vision-Language reasoning with key strategies for Unsupervised Domain Adaptation (UDA). First, we improve the fine-grained segmentation capabilities of VLMs through multi-scale contextual data, robust text embeddings with prompt augmentation, and layer-wise fine-tuning in our proposed Foundational-Retaining Open Vocabulary Semantic Segmentation (FROVSS) framework. Next, we incorporate these enhancements into a UDA framework by employing distillation to stabilize training and cross-domain mixed sampling to boost adaptability without compromising generalization. The resulting UDA-FROVSS framework is the first UDA approach to effectively adapt across domains without requiring shared categories.
\end{abstract}
\section{Introduction}
\label{sec:intro}

Semantic segmentation, the task of assigning a categorical label to every pixel in an image, is critical for multiple applications. However, its class dviersity is constrained by the number and nature of the annotated classes learned during training. To handle this limitation, two approaches have been explored: (1) training in synthetic domains where classes can be created at will and instances are annotated at creation; so-trained models need to be later adapted to the target domain. (2) leverage the foundational encoded knowledge of Vision Language Models (VLMs) by adapting their outcomes to a segmentation setting. 

First, Unsupervised Domain Adaptation (UDA)  \cite{wang2024class, AlcoverOnExploring, ma2024taming} aims at leveraging synthetic datasets, where simulated environments enable the automatic generation of pixel-perfect annotations at a fraction of the cost of manual labeling \cite{cheng2024learning,liang2024adaptive, zhou2024multi3d}. While UDA has shown remarkable success in mitigating domain shifts between tasks and data domains, UDA has a significant drawback: its limited ability to adapt to new categories absent in the synthetic dataset \cite{ Alcover-Couso_2023_ICCV}. A potential solution would be to regenerate the synthetic data with the new categories, but the models would have to be re-trained, as UDA models are notoriously known for overspecificity \cite{li2024cross}. This limitation impairs the performance of UDA-based models in real-world applications where coping with unseen categories is crucial. 

\begin{figure}
	\centering
	\includegraphics[width=\linewidth]{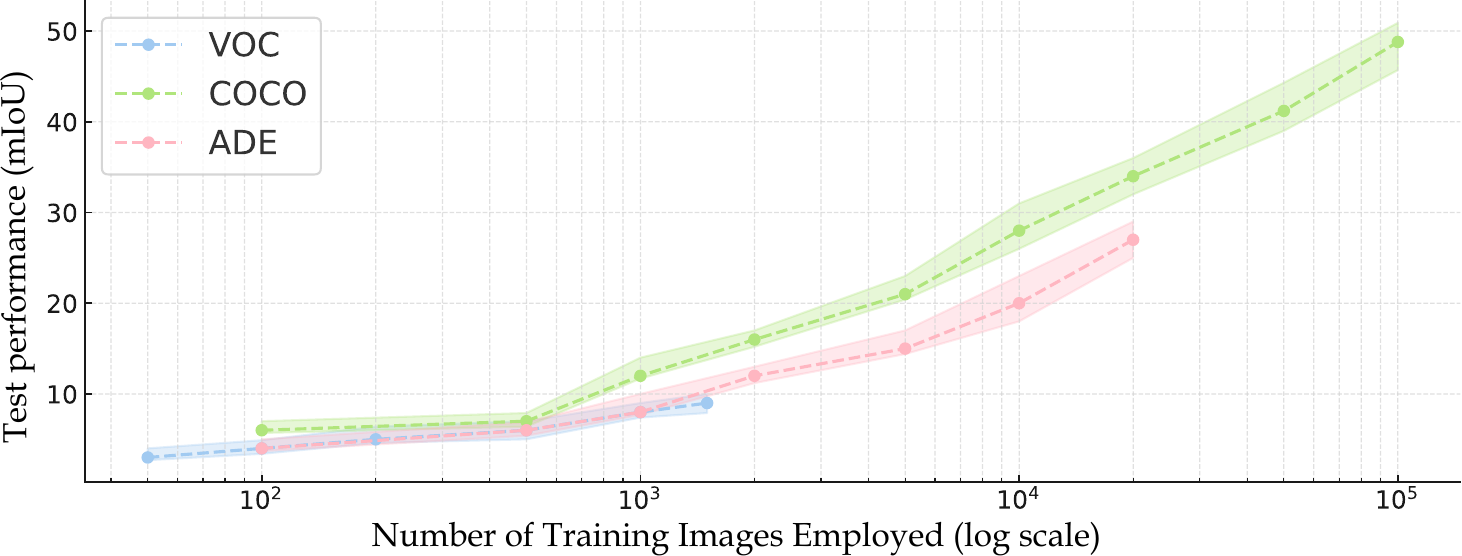}
	\caption{\textbf{State of the art for open vocabulary semantic segmentation underperforms when trained with small training sets}. Results of CAT-Seg \cite{cho2023catseg} trained on random subsets derived from three popular datasets with different amounts of images across three random seeds (maximum minimum range depicted by shadowed area). Performance evaluated in the COCO validation set \cite{coco}.} 
	\label{fig1}
\end{figure}

Second, recent advancements in open vocabulary semantic segmentation (OVSS)   have shown that VLMs can be adapted to perform dense prediction tasks. However, these methods typically require task-specific data for effective fine-grained adaptation to semantic segmentation tasks (see Figure \ref{fig1}). Unfourtunately, during this adaptation process,generally conducted by fine-tuning, the foundational capabilities of the model are catastrophically \cite{Ding2022Don't, Yan2022Generative}. The practical applicability of these model is then limitted, as it is still to appear a technique to enable adaptation without reducing the generality of the prior features \cite{zhou2022maskclip, Srinivasan2022CLiMB}

While these challenges have been tackled separately, in this paper we show that their solutions can be mutually reinforcing.

Our key insight is two-fold: First, UDA principles and techniques can dramatically reduce the data requirements for the fine-grained adaptation of VLM by providing efficient mechanisms for employing unlabelled data and preserving the global knowledge from VLM pre-training. Second, the rich semantic understanding and open vocabulary capabilities of VLMs can help UDA methods evolve from their closed-set constraints, enabling them to handle novel categories through zero-shot recognition capabilities.

\begin{figure*}[tp]
	\centering
	\begin{subfigure}[b]{.32\linewidth}
		\includegraphics[width=\linewidth]{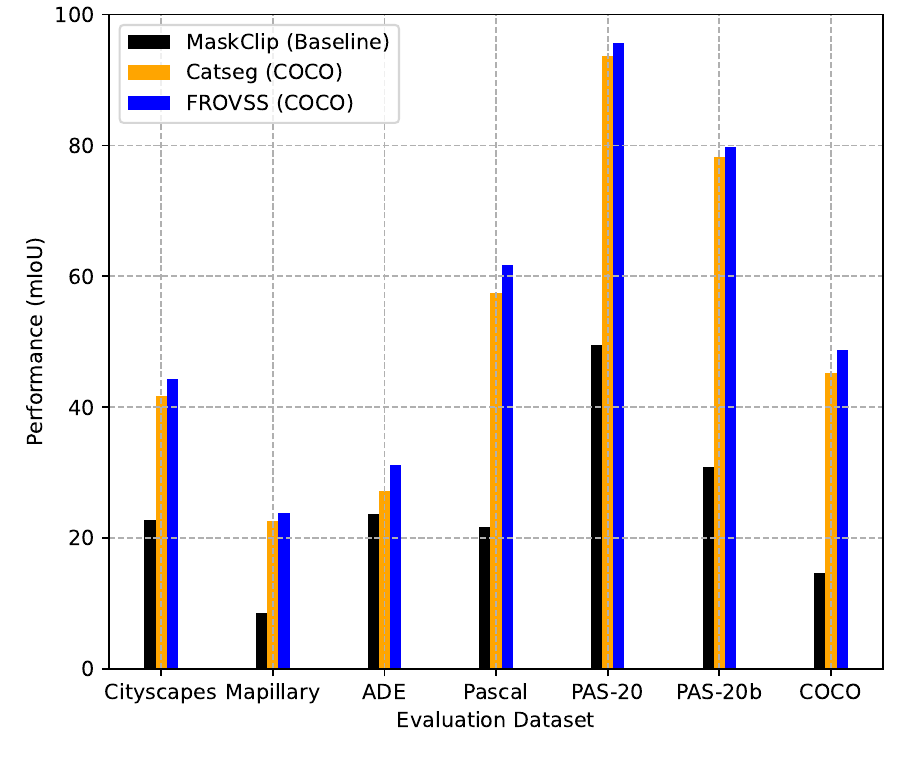}
		\caption{Our proposals model improve SOTA models for the default setup of COCO training.\\}
		\label{fig2a}
	\end{subfigure}\hfill
	\begin{subfigure}[b]{.32\linewidth}
		\includegraphics[width=\linewidth]{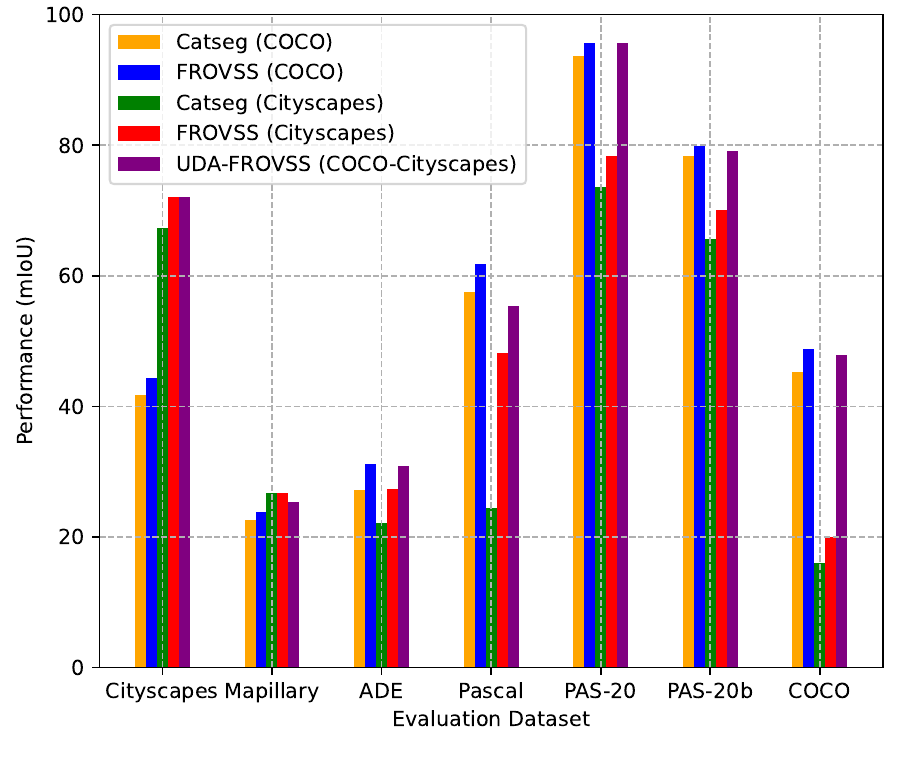}
		\caption{Urban scenes datasets improve specificity at the cost of reduced generalization (see Cityscapes and COCO performance)}
		\label{fig2b}
	\end{subfigure}\hfill
	\begin{subfigure}[b]{.32\linewidth}
		\includegraphics[width=\linewidth]{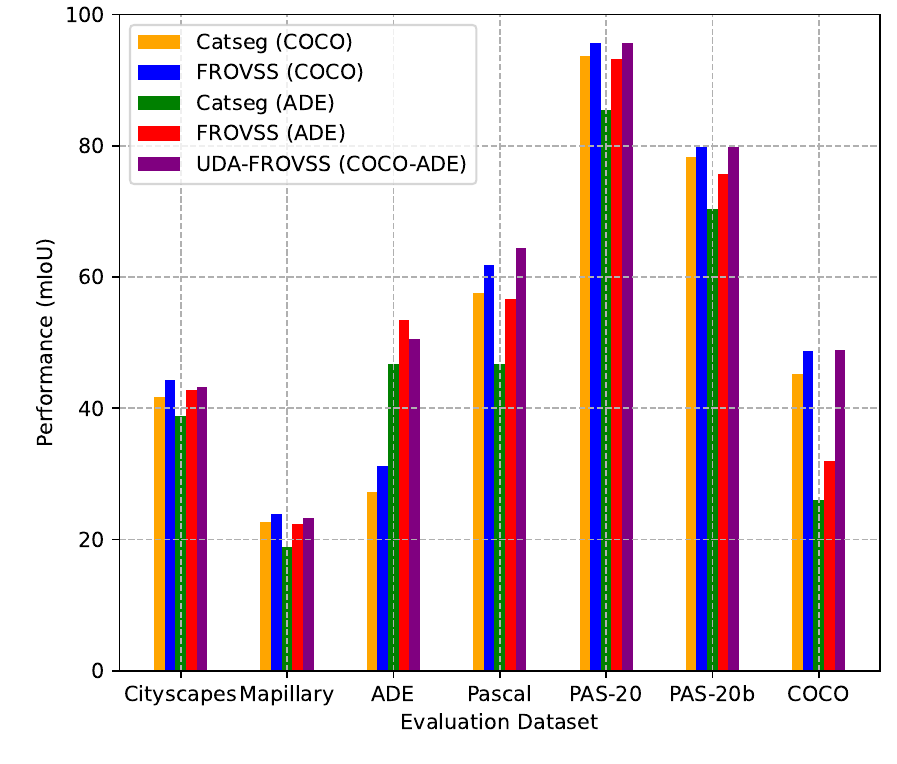}
		\caption{Scene parsing datasets improve specificity at the cost of reduced generalization (see ADE and COCO performance)}
		\label{fig2c}
	\end{subfigure}
	\caption{\textbf{Visual summary of contributions}. In Figure \ref{fig2a}, we showcase the benefits of FROVSS in the standard OVSS setup (trained in the COCO dataset and evaluated in multiple datasets). Figures \ref{fig2b} and \ref{fig2c} illustrate the major challenge we tackle: training with task-specific datasets (Cityscapes in \ref{fig2b} and ADE in \ref{fig2c}) drastically reduces generalization of the model. To overcome such issue, our proposed combination of UDA and OVSS (UDA-FROVSS), presents high performance for task-specific datasets while preserving generalization across other datasets (see UDA-FROVSS in \ref{fig2b} and \ref{fig2c}). Note that these UDA models do not require labels for the task-specific dataset. } 
	\label{fig2}
\end{figure*}

Based on these challenges, our contributions can be organized into three main areas:

\begin{itemize}
	\item \textbf{Open Vocabulary Model Enhancements}: We introduce a novel decoder architecture that leverages convolutional layers  for aiding transformer layers learning with limited data \cite{NEURIPS2021_ff1418e8, Yuan_2021_ICCV,ZHANG2023109020}. Additionally, our fine-tuning strategy is designed to preserve the integrity of VLM's pre-trained weights, avoiding catastrophic forgetting while enhancing pixel-level predictions (see Figure \ref{fig2a}). 
	\item \textbf{Textual Relationship Improvements}:  To improve the cross-dataset generalization ability of our model, we propose a concept-level prompt augmentation strategy. By using Large Language Models (LLMs) and providing specific instructions for annotators, we generate diverse and contextually enriched textual prompts. Our approach enhances the model's ability to recognize categories through semantic relationships across datasets (see comparison with \cite{cho2023catseg} in Figure \ref{fig2}  Foundational-Retaining Open Vocabulary Semantic  Segmentation (FROVSS)).
	\item \textbf{Synergy between UDA and Open Vocabulary}: We bridge the gap between UDA and OVSS by developing a unified framework that eliminates the need for shared categories between the source and target domains, making our UDA framework the first to enable models to recognize and segment objects beyond the categories encountered during training. By combining the domain generalization capabilities of UDA with the flexibility of OVSS, we train models which present  high effectiveness on the target dataset while preserving their generalization prowers (see Figures \ref{fig2b} and \ref{fig2c} UDA-FROVSS).
\end{itemize}

Alltogether, these contributions enable open vocabulary models to significantly benefit from the use of unlabeled images common in UDA, thus enhancing their performance and generalization across diverse datasets.   As a result, the models trained using FROS and UDA-FROS demonstrate superior segmentation accuracy on the training dataset but also enhanced transferred performance in previously unseen domains (see Figure \ref{fig2}). 

The proposed methods are extensively evaluated across multiple semantic segmentation datasets \textbf{showcasing improvements across all analyzed benchmarks}: PAS-20 ($\uparrow 2.0\%$ mIoU), COCO ($\uparrow 3.2\%$ mIoU), ADE-20 ($\uparrow 7.9\%$ mIoU),  Pascal ($\uparrow 17.2\%$ mIoU) and Cityscapes ($\uparrow 22.1\%$ mIoU). On top of them, we also establish a new benchmark for UDA in the Synthia-to-Cityscapes setup, surpassing the previous state-of-the-art frameworks by over  $8\%$ in mIoU.


The rest of the paper is organized as follows: Section \ref{sec:RW} reviews related work on OVSS and UDA; Section \ref{sec:Method} details the proposed OVSS architecture, prompt definition, and UDA training framework;  Section \ref{sec:Exp} covers experimental results, focusing on both generalization with a single dataset and specialization to an unlabeled set (UDA); Finally, Section \ref{sec:Cc} discusses insights from the experiments and proposes future research directions for OVSS.

\section{Related Work}
\label{sec:RW}

\paragraph{Open Vocabulary Semantic Segmentation}
Early CLIP \cite{CLIP} extensions to pixel-level prediction primarily involved its use as an image classifier with masked regions \cite{Luo2023SegCLIP, Ding2022Open-Vocabulary, Huynh2021Open-Vocabulary, benigmim2023collaborating, Xu2023side, li2024clip}. These methods typically utilized a region proposal network to identify image segments, which were masked and classified using a static CLIP model. This process required processing each segment through the CLIP encoder, making it computationally intensive. Recent studies \cite{li2022exploring, li2023clip, cho2023catseg, hoyer2023semivl, zhou2022maskclip} have shifted focus to harnessing CLIP's features for more detailed, object-level semantic analysis. Initial efforts involved using CLIP's attention maps to create segmentation maps \cite{li2022exploring, li2023clip}, which were classified based on similarity values. However, as CLIP is trained for global image representations, these methods struggled to capture intricate image details. 
Alternative approaches \cite{liang2023open,9879044,xu2021} introduce category-agnostic region proposals as support inputs to the CLIP encoder. This process has the limitation of losing the image global context as each support region is processed independently.  In contrast, MaskCLIP \cite{zhou2022maskclip} addressed this by modifying CLIP's final pooling layer to extract dense features directly. To obtain accurate dense features, they fine-tuned the query and key embeddings responsible for spatial representation. Building on this, CAT-Seg \cite{cho2023catseg} proposes to rely on similarity maps between these dense visual features and textual category descriptions as inputs for a decoder, thereby avoiding the direct CLIP optimization that may be compromising the open vocabulary capabilities of the image backbone \cite{zhou2022maskclip, Srinivasan2022CLiMB}.  While the CAT-Seg method is proved to be effective for transferring the open vocabulary capabilities of CLIP to the segmentation task, but the added decoder and the modified layers of CLIP are domain-specific. Consequently, it underperforms in scenarios subjected to semantic distribution shifts. To address this, we first propose defining robust text embeddings by incorporating synonyms and annotator instructions to refine and focus the semantic meaning of each category name within each dataset. Second, we introduce a layer-wise learning rate to enhance the fine-tuning of VLMs, mitigating the catastrophic forgetting. Finally, we design a decoder that combines convolutional and transformer layers, enabling efficient adaptation of VLMs to dense prediction tasks. Together, these techniques form the foundation of our FROVSS method.  

\paragraph{Unsupervised Domain Adaptation}
Addressing the challenge of applying open vocabulary segmentation to unseen datasets, we advocate the use of UDA techniques \cite{MTAP,10128983,PCCL}. UDA leverages knowledge from a source-labeled domain to train models that can effectively generalize to unlabeled target domains by coping with a covariant distribution shift. To that end, pseudo-labels \cite{Lee2013PseudoLabelT} from a teacher model are used to guide the learning of a model being trained (i.e., the student) \cite{hoyer2022daformer,hoyer2022hrda,CDAC,kumar2023conmix,Alcover-Couso_2023_ICCV,hoyer2023mic}. As UDA frameworks typically lack a reliable teacher, the common choice is to define the teacher model as an exponential moving average of the student's weights. This allows integrating learnt knowledge on the fly, while mitigating concept drifting by hampering the impact of pseudo-labels in the teacher model \cite{Tarvainen2017MeanTA}. Concept drift is the phenomenon where accumulating inaccuracies, particularly false positives, misguides the model's learning trajectory \cite{Guo2022Research,Zheng2020Rectifying}. Among other canonical UDA strategies, cross-domain mixed sampling (image mixup) randomly overlays images from both domains to enforce the model to learn domain-invariant features \cite{Yamada1989Characterizations,Tremblay2018Training,Prakash2018Structured,Valtchev2020Domain,Tranheden2020DACSDA}. Moreover, domain randomization introduces controlled variability in the training data, such as changes in lighting, textures, and other environmental factors. Additionally, it proposes overlaying source objects on target images to further introduce variability \cite{Tranheden2020DACSDA, Tremblay2018Training}. These variations ensure that the model remains adaptable even when faced with data different from the training set. We blend these UDA techniques into our VLM-based OVSS method: UDA-FROVSS, yielding large performance gains across all analyzed domains.

\paragraph{Robust Text Embeddings} To improve open vocabulary capabilities, state-of-the-art methods \cite{CLIP,ma2024clip} employ multiple descriptions of the image to generate a mean embedding. These mean embeddings are supposed to be more reliable descriptors of the object as the only common factor of the sentences is the text identifying the target object, therefore reducing the noise introduced into the text embedding of the prompt. Differently, the state-of-the-art method for open vocabulary segmentation \cite{cho2023catseg} does not employ mean text embeddings and computes similarity on each prompt, thus neglecting the cascaded benefits of better textual representations. 
To overcome this drawback we employ different prompt augmentation techniques based on object characteristics, e.g. ``A photo of a \textit{Vintage} car'', and descriptions for annotators. 


\section{Method}
\label{sec:Method}

\begin{figure*}[t]
	\centering
	\includegraphics[width=\linewidth]{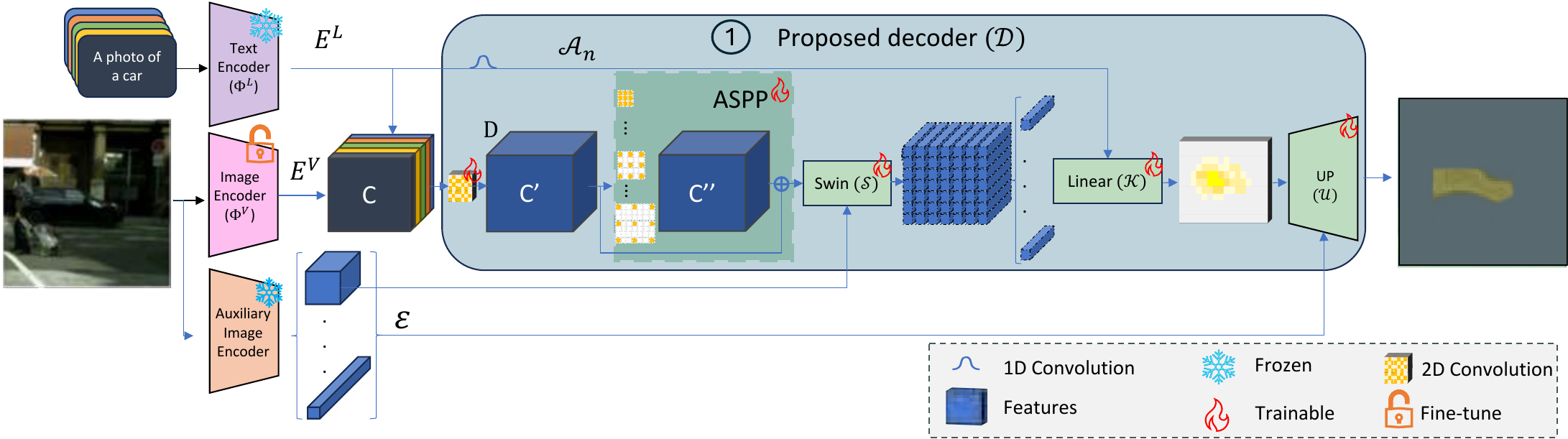}
	\caption{\textbf{Proposed decoder for open vocabulary semantic segmentation}, exemplified with the category: ``car''. We guide segmentation by refining the similarities between dense features extracted from the image encoder and the text features ($C$).}
	\label{fig:Architecture}
\end{figure*}

In this paper, we enhance the adaptability of CLIP to interpret semantics at pixel level across novel domains. We initialize the training with a pre-trained VLM as it contains rich semantics learned from large-scale training. We follow \cite{zhou2022maskclip} and modify the last layer of the image encoder to obtain dense image features which are combined with text embeddings to generate cost volume embeddings as in \cite{cho2023catseg}. These cost volumes are then fed to a decoder to generate segmentation maps (see Figure \ref{fig:Architecture}).  Additionally, to improve performance we propose \textit{paraphrasing ensemble} to generate robust text embeddings, these robust text embeddings can be applied during the training or the testing phase to improve performance. 

\subsection{Open Vocabulary Semantic Segmentation}\label{sec:proposed-ovss}
\paragraph{Problem overview}
In open vocabulary semantic segmentation, the task is to map each pixel of an input image $x$ to a label from a set of $N$ potential categories $\mathcal{P} =\{P_1,...,P_n,...,P_N\}$, where for each $n$-th category, $P_n = \{p_{1,n},...,p_{m,n},...,p_{M,n}\}$ is a set of $M$ different text prompts $p_{m,n}$ (equal $M$ for all categories). This task expands upon traditional semantic segmentation by being able to cope with unseen categories during inference, posing a challenge beyond conventional segmentation capabilities.

\paragraph{Fine-Tuning of CLIP}
Prior works utilizing CLIP \cite{zhou2022maskclip, cho2023catseg, Ding2022Don't, Yan2022Generative, Srinivasan2022CLiMB} highlight that conventional fine-tuning of the image encoder can degrade performance due to potential misalignment of the image and text encoders. Therefore, guided by the insight that tuning layers responsible for spatial interaction (e.g., attention layers and positional embeddings) suffices for transferring image-level representations to pixel-level \cite{NIPS2017_3f5ee243}, we freeze the MLP layers in the encoder. Additionally, we follow the hypothesis that deeper layers in the image encoder encapsulate task-specific filters, while shallow layers represent task-agnostic filters which should be less tuned to obtain optimal performance \cite{FinetuningTC, Wu2018Fast}. Therefore, we propose to decrease the learning rate in each layer by a factor of $\beta$ with respect of  the previous layer:
\begin{equation}
	lr_{l} \leftarrow lr_{l+1}\cdot \beta,
	\label{eq:finetuning}
\end{equation}
where $lr_l$ is the learning rate assigned to layer $l$ and $\beta$ a training hyper-parameter. For the framework we only set the initial learning rate of the last layer of the encoder and then the learning rate is propagated to the following layers.

\paragraph{Language-Guided Cost Volume and Semantic Decoding}
The vision-language pre-training allows for aligned vision and language embeddings; such alignment enables the reasoning of semantics of the image given a description. We employ a decoder to disentagle the relationships derived from the VLM into accurate pixel-level predictions. To do so, first we modify the CLIP image encoder as per \cite{zhou2022maskclip}, so that it generates patch-level image features by removing the attention pooling of the last layer of the ViT encoder \cite{dosovitskiy2021an}. 
For each input image $x$ of $H \times W$ spatial resolution, the modified CLIP image encoder $\Phi^V(x)$ processes image patches of size  $k \times k$ to obtain sets of visual features $\{E^V_i, i \in [1,H' \times W']\}$, with $H' = \frac{H}{k}$ and $W' = \frac{W}{k}$. Regarding the text branch, the encoder $\Phi^L(\mathcal{P})$ is kept unaltered, yielding a set of text features $E^L_{m,n}$ for each text description and category of the same size as $E^V_i$. 

We define a cost volume embedding $C\in\mathbb{R}^{(H'\times W')\times M\times N}$ as the cosine similarity \cite{Rocco17} between $E^V_i$ and $E^L_{m,n}$:

\begin{equation}
	C_{i,m,n} = \frac{E^V_i \cdot E^L_{m,n}}{\lVert E^V_i\lVert \lVert E^L_{m,n}\lVert}.
	\label{eq:similarity}
\end{equation}

These cost volumes are fed into a decoder $\mathcal{D}$, whose main objective is to refine the similarities extracted from the text and image encoders. Our decoder is composed of a spatial refinement module followed by a semantic reasoning module. Initially, each  category-prompt-spatial feature $C_{:,:,n}$ is embedded into a  $D$ hidden dimension  by means of a a convolution yielding $C'\in\mathbb{R}^{(H'\times W')\times D \times N}$.

To aid in the spatial reasoning, we incorporate a residual ASPP module to train long-range and multi-scale context relations across all spatial regions of these similarity maps $C'_{:,:,n}$. This module incorporates these relationships yielding $C'' \in\mathbb{R}^{(H'\times W')\times D \times N}$.

\paragraph{Visual Guidance Branch}
Our framework may incorporate auxiliary image feature embeddings $\mathcal{E}$ for spatial structure or contextual information through additional image encoders. 
These embeddings are concatenated with the spatially-refined cost volume features and processed by two Swin Transformer blocks $\mathcal{S}$:
\begin{equation}
	F_{:,d,n} = \mathcal{S}([C''_{:,d,n}; \mathcal{E}]),  \text{ } d \in [1,D].
\end{equation}
The Swin Transformer blocks combine spatial information from the auxiliary encoder with relational features from the cost volumes, enabling the model to fill gaps and refine semantic structures from the VLM based on spatially-aware context.

\paragraph{Text Guidance Branch}
To reinforce the textual guidance, we incorporate a semantic reasoning module that reinforces the relationships among the textual descriptions in $F$ without further modifying the spatial ones. Through a linear kernel $\mathcal{K}$, we leverage our prompt-augmented semantic anchors $\mathcal{A}_n$ (see section \ref{sec:Prompt}) obtained as a linear combination of the $M$ text embeddings of each category. Empirically, we set $\mathcal{K}$ as a linear transformer:
\begin{equation}
	F'_{i,:,n} = \mathcal{K}([F_{i,:,n} ; \mathcal{A}_n]).
	\label{eq:linear}
\end{equation}

\paragraph{Up-Sampling Module}

Given that the underlying CLIP visual transformer encoder operates on a $k \times k$ times smaller feature resolution than the input, the similarity volume $F'$ is up-sampled by a module $\mathcal{U}$ to recover the original image resolution. This module is the concatenation of bilinear up-sampling followed by a set of transposed convolutions. This process iterates as many times required \footnote{For instance for $P=16$ two iterations are required.} to yield an output of the same resolution as the input image for each hidden dimension and category: $F'' \in \mathbb{R}^{(H \times W)\times D \times N}$. 

\begin{equation}
	F''_{:,d,n} = \mathcal{U}(F'_{:,d,n}) \label{eq:upsampling}
\end{equation}

Incorporating visual guidance \cite{10.1007/978-3-031-19818-2_7}, the auxiliary feature embeddings $\mathcal{E}$ are concatenated with $F'$ and processed by a convolutional layer to return to the hidden dimension $D$ before the up-sampling module. These features are concatenated and processed in the same order they are extracted. Intuitively, following the extraction order in the auxiliary decoder refines finer details of the image. 
Finally, each category's similarity $\mathcal{F}_{:,n} \in \mathbb{R}^{(H \times W) \times N} $ is computed for each spatial position by the sigmoid activation of a weighted combination of the hidden dimension $D$ in $F''$, implemented as a learnable $1 \times 1$ convolution.

\subsection{Prompt Definition}
\label{sec:Prompt}
\begin{figure*}[t]
	\centering
	\begin{subfigure}[b]{.5\linewidth}
        \vfill
		\centering
		\includegraphics[width=\linewidth]{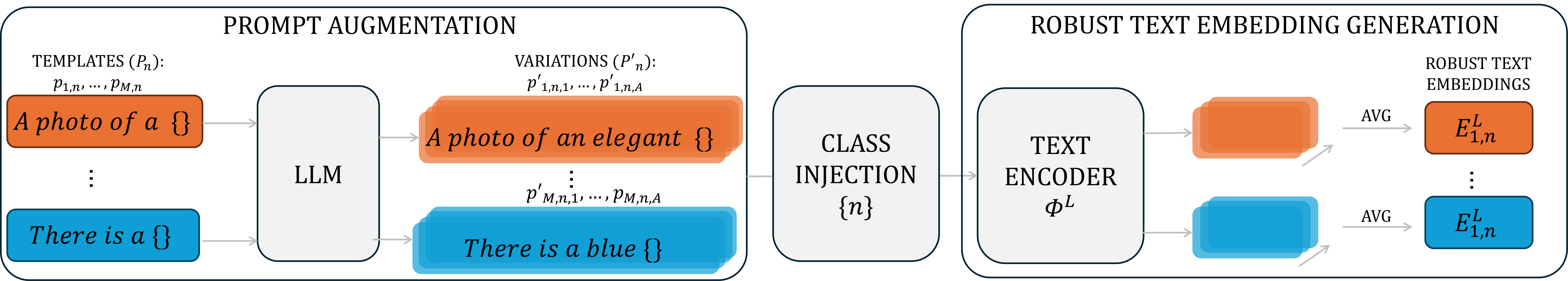}
		\caption{Robust per-prompt embeddings}
		\label{subfig:robust}
        \vfill
	\end{subfigure}\hfill
	\begin{subfigure}[b]{.5\linewidth}
		\centering
		\includegraphics[width=\linewidth]{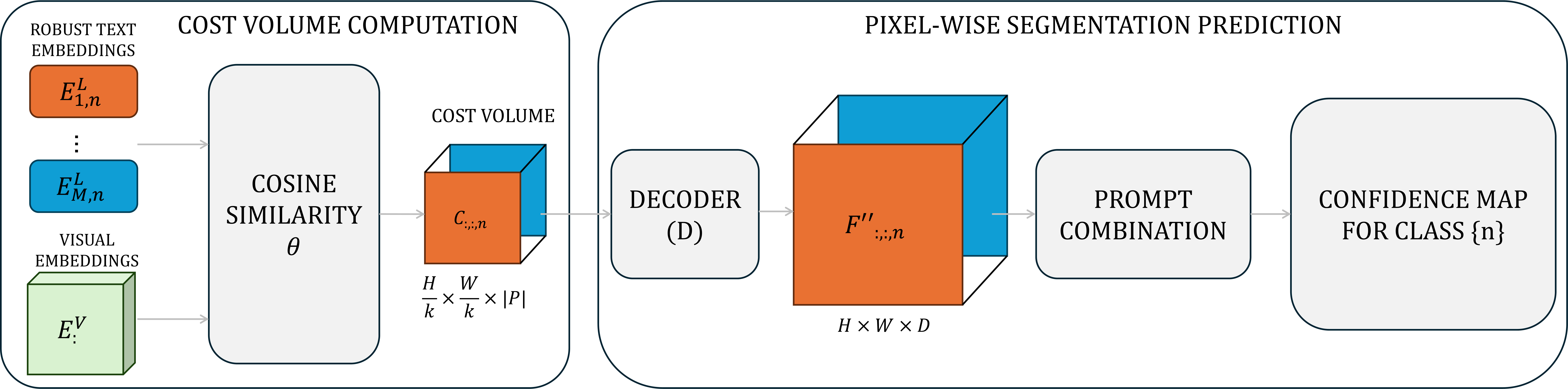}
		\caption{Prompt unification}
		\label{subfig:unification}
	\end{subfigure}
	\caption{\textbf{Prompt augmentation pipeline.} }
	\label{fig:prompts}
\end{figure*}

We find two major drawbacks in prompt definition of the current CLIP-based segmentators. First, they do not employ mean text embeddings nor prompt augmentations to generate more reliable text embeddings \cite{cho2023catseg}. Second, they do not account for conflicting category names given by different datasets, e.g. the \textit{Cityscapes} dataset \cite{Cordts2016Cityscapes} differentiates between the categories \textit{rider} and \textit{pedestrian} , whereas other datasets group both under the category \textit{person}, without accounting for the situational context of the individual.

To address the identified issues, we incorporate the descriptions provided to annotators into the text prompts for each category. These descriptions offer additional detail on how each category is defined in the target dataset, enhancing specificity without compromising generalization to other datasets. Additionally, we use LLMs \cite{brown2020language} to generate synonyms for each category to improve concept robustness. Duplicate synonyms are removed to maintain category distinction. Furthermore, we use LLM-generated variations to create multiple versions of each prompt (see Figure \ref{subfig:robust}). This approach enables a novel prompt-level augmentation protocol, leveraging robust text embeddings for each prompt instead of averaging across prompts. We propose to define augmentations based on:

\begin{itemize}
	\item \textbf{Object characteristics augmentation}. These augmentations are based on including different adjectives before the class name into the prompt, e.g.  ``A photo of a \textit{Vintage} car''. These object characteristics should provide robustness to the concept as these variations are averaged with the concept as a common anchor.
	\item \textbf{Photometry of the image augmentation}. At the end of the prompt followed by a comma we include visual characteristics of the global image, e.g. ``A photo of a car, \textit{with High Contrast}''. These characteristics may be useful to provide robustness towards style changes.
	\item\textbf{ Background characteristics augmentation}. We include positional information into the prompt, e.g. ``A photo of a car \textit{in the Countryside}''. These characteristics can be useful for extrapolating from datasets captured on specific geographical points, such as Cityscapes captured on Germany \cite{Cordts2016Cityscapes}, to more diverse datasets as Mapilliary \cite{MVD2017} captured on a global scale.
\end{itemize}

Formally, we propose to generate $A$ variations of each category textual description $p_{m,n}$, generating an augmented set  $P'_{n} = \{p'_{1,n,1},...,p'_{1,n,A}, ..., p'_{m,n,A}\}$. Then, these are fed to the text encoder to extract text features. Before combining them with the visual ones in Equations \ref{eq:similarity} and \ref{eq:linear}, we calculate the mean of the augmented set of text features to obtain a robuster set of text features: 
\begin{equation}
	E^L_{m,n} = \sum_{a=1}^A\frac{\Phi^L(p'_{m,n,a})}{A}.
\end{equation}

To unify the responses across the different augmentations, a $1 \times 1$ convolution is trained, thereby relying on a learnable weighted combination (see Figure \ref{subfig:unification}). 




\subsection{Unsupervised Domain Adaptation} \label{uda}
\begin{figure*}[t]
	\centering
	\includegraphics[width=\linewidth]{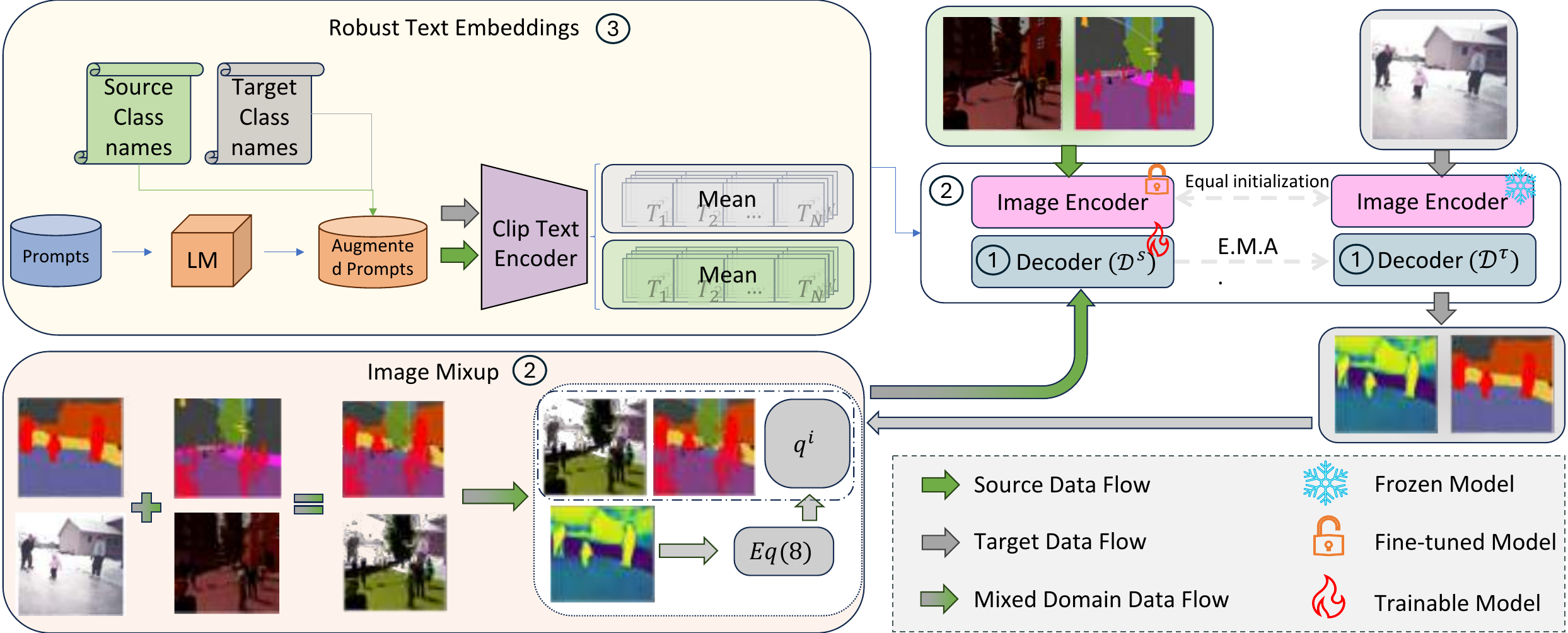}
	\caption{\textbf{Overview of UDA-FROVSS, which combinines VLMs with UDA}. Key Components are illustrated within delineated boxes: (1) Integration of a custom decoder alongside a fine-tuning strategy to effectively train the framework; (2) Adaptation of UDA techniques, incorporating a teacher-student framework and image mixup for domain robustness; (3) Generation of robust text embeddings for enhanced category recognition}
	\label{fig:Framework}
\end{figure*}
 Our UDA-FROVSS framework, illustrated in Figure \ref{fig:Framework}, integrates the improved CLIP capabilities in the UDA framework to utilize labeled images from the source domain to guide the learning on unlabeled images on the target domain. While UDA already achieves good segmentation quality on shared categories, current UDA frameworks struggle to segment target-private categories not present in the source domain. To address this, we propose to combine open vocabulary segmentation with UDA techniques into our UDA-FROVSS framework. Specifically, we adapt UDAs teacher-student training scheme and image mixup to enhance domain generalizatio by employing pseudo-labels generated by the teacher on the target dataset. Moreover, to preserve the open vocabulary capabilities of the final model, we propose to only update the teacher decoder, as the backbones update leads to concept drifting, thus corrupting the open vocabulary reasoning outside the vocabulary employed during training. Therefore, the teacher preserves the open recognition at the cost of fine-grained classification, as the encoder remains unaltered.

\paragraph{Problem Overview}
In the context of domain adaptation, a model is trained on labeled source data and unlabeled target data. Training a model with source data typically leads to suboptimal performance when applied to target domain images. For the fine-grained adaptation of VLMs to OVSS, one must adapt the model to transfer the learned image-level representations to pixel-level ones, leading to domain specificity of the learned layers \cite{zhou2022maskclip, cho2023catseg}. 

\paragraph{Teacher-Student Framework}
Our proposal employs a teacher-student framework from two equally initialized VLMs, each one composed of a decoder ($\mathcal{D}^{\tau}$ and  $\mathcal{D}^{s}$ for the teacher and student respectively) and an encoder. The student model $\mathcal{M}^{s}$ learns through cross-entropy loss on the source data using each source image $x$ with one-hot encoded label $y$:

\begin{equation}
	\mathcal{L}(x) = -\sum_i^{H\times W} y_i log(\mathcal{M}^s(x)_i), 
\end{equation}

\paragraph{Image Mix-Up}
Our hypothesis is based on CLIP inherently supporting open vocabulary for broadly defined categories, so our focus is on training the head to discern shapes and align with CLIP's knowledge at a more detailed, fine-grained level. Therefore we follow a domain randomization protocol \cite{Tobin2017DomainRF, Tremblay2018Training, Tranheden2020DACSDA} to improve the shape understanding of the model. Following \cite{Tranheden2020DACSDA}, we overlay half of the semantic instances in the source domain on top of target images to generate blend data-augmented images: $x'$. The blending is also performed to generate the labels $y'$ by combining the ground truth labels with the pseudo-labels. To cope with pseudo-labels uncertainty, we employ a weighted cross-entropy loss for mix-up training:
\begin{equation}
	\mathcal{L'}(x') = -q\sum_i^{H\times W}y'_i log(\mathcal{M}^s(x')_i),
\end{equation} 
where $q$ is the quality of the target image segmentation; $q$ is computed through the confidence of the pixel-wise predictions of the teacher model by using a pixel-wise confidence threshold ($\mu$):
\begin{equation}
	q = \frac{\sum_i^{H\times W} max(\hat{y}_{i,:}) > \mu}{H \cdot W},
\end{equation}

\paragraph{Teacher Update}
Based on observations that updating the teacher encoder improves performance on source domain but diminishes the model's open vocabulary potential, we propose to keep the encoder of the teacher model unaltered, whereas the teacher decoder is updated through an exponential moving average (E.M.A.) of the student weights implementing a temporal ensemble at every time step $\delta$ to stabilize predictions:
\begin{equation}
	\mathcal{D}^{\tau}_{\delta+1} \leftarrow  \alpha \cdot \mathcal{D}^{\tau}_\delta + (1-\alpha)\cdot \mathcal{D}^s_\delta.
\end{equation}


As the teacher decoder becomes increasingly unaligned with its encoder, we propose to progressively incorporate the student predictions in the pseudo-labels. Our aim is to first only weight teacher predictions in the learning,  to seamlessly employ the student predictions once its learning stablizes.  We propose to implement this as a weighted linear combination of the teacher ($\mathcal{M}^{\tau}$) and student predictions:
\begin{equation}
	\hat{y}_i = \gamma\cdot \mathcal{M^{\tau}}(x) + (1-\gamma)\cdot \mathcal{M}^s(x),
\end{equation}

Initially, generated pseudo-labels  only takes into account the teacher pseudo-label, as it has been shown that CLIP possesses somewhat reliable zero-shot segmentation capabilities \cite{zhou2022maskclip}. To do so, we define $\gamma$ as a parameter reduced at every time-step:

\begin{equation}
	\gamma_{\delta+1} \leftarrow (\frac{1}{\delta^{\gamma_{\delta}}+1}),
\end{equation}

where $\gamma_0$ is a training hyper-parameter. We select this update, as it is a smoothed version of the teacher decoder E.M.A. update. Therefore, the more the teacher decoder changes the less we trust the its pseudo-labels as its decoder will be increasingly unaligned with its  encoder.

\section{Experiments}
\label{sec:Exp}
\subsection{Experimental Setup}
\paragraph{Datasets and Evaluation}
\begin{table}[t]
	\centering
	\setlength{\tabcolsep}{3pt} 
	\resizebox{1\linewidth}{!}{%
		\begin{tabular}{l c c c c c c c}
			Dataset & COCO& CS &MAP & ADE-20&PC& PAS& Synthia \\\toprule\midrule
			Categories& 171& 19& 65& 150& 59& 20 & 16\\
			Train& 118& 3& 18& 20& 5& 1.5&9.5\\
			Test&5&0.5&2&2&5&1.5& -\\\bottomrule
		\end{tabular}}
	\caption{\textbf{Summary of the number of defined categories and subset sizes (in thousands) for the datasets used in the following experiments.} ``-'' denotes the absence of a defined test set.}
	\label{tab:datasets}
\end{table}
Our experiments explores seven distinct datasets for semantic segmentation, each used independently: COCO-stuff \cite{coco},  Cityscapes (CS) \cite{Cordts2016Cityscapes}, Mapillary (MAP) \cite{MVD2017}, ADE-20 \cite{zhou2019semantic}, Pascal-Context (PC) \cite{mottaghi_cvpr14}, Pascal VOC \cite{Pascal} and Synthia \cite{Ros2016}. Synthia is the only synthetic dataset. Cityscapes, Mapillary and Synthia are urban scenes datasets while the others are general-purpose ones. For the  Pascal VOC dataset, we present results in two formats: including the background (PAS$^b$), and focusing solely on object categories (PAS). Table \ref{tab:datasets} provides further details.   

As evaluation metric, we adopt the per-class  mean Intersection over Union at pixel-level, $mIoU=\frac{TP}{TP+FN+FP}$, where TP, FP and FN stand for true positives, false positives and false negatives respectively.

\paragraph{Implementation details}
Our models are trained on a single GPU A40 with a per-pixel binary cross-entropy loss, batch size of 4, AdamW optimizer with a learning rate of $2\cdot 10^{-4}$ for the decoder and $2\cdot 10^{-6}$ for the CLIP image encoder, with weight decay of $10^{-4}$. The CLIP text encoder always remains frozen. Our models utilize ViT-B \cite{dosovitskiy2021an} as the CLIP image encoder ($P=16$), and a Swin Trasformer \cite{liu2021swinv2} as the auxiliary image encoder. The image encoder remains frozen when initialized from pre-trained weights, if not, the learning rate is set to $2\cdot 10^{-4}$. All models are trained for 80k iterations. We set $\mu=0.96$ and $\beta=0.95$ through initial exploration and $D=128$ following \cite{cho2023catseg}.

\subsection{Results on Open Vocabulary Segmentation}
This subsection provides results for FROVSS without using any UDA techniques (see Section \ref{sec:proposed-ovss} and Figure \ref{fig:Architecture}), following standard protocols for comparisons.

\begin{table}[]
	\centering
	\setlength{\tabcolsep}{6pt}
	\resizebox{1\linewidth}{!}{%
		\begin{tabular}{l|ccccc}
			Method& CS &MAP & ADE-20&PC&COCO\\\toprule\midrule
			CAT-Seg \cite{cho2023catseg}&67.3*&52.6*&46.8&62.4&45.3\\
			\rowcolor{Gray}
			Proposed & 72.1 &53.4&53.4&67.4&48.8\\         
			\bottomrule
	\end{tabular}}
	\caption{\textbf{Decoder performance comparison against the best stage-of-the-art for open vocabulary semantic segmentation.} Training and test data correspond to the same dataset. Not reported results are trained by us and indicated by: ``*''.}
	\label{tab:architectural_res}
\end{table}
\paragraph{Open Vocabulary Segmentation Architecture.}
Table \ref{tab:architectural_res} compiles the results of our baseline decoder and compares them to the CAT-Seg architecture. Aditionally, Table \ref{tab:architectural_res_complete} presents the full performance comparison with the baseline framework CAT-Seg \cite{cho2023catseg}. 

Across all analyzed training and validation setups, our decoder improves or performs on-par with CAT-seg. These improvements are specially notorious on dense labelled datasets such as Cityscapes and Mapillary, suggesting that our proposed encoder does in fact improve segmentation of fine details and densly populated scenes. 

\begin{table}[]
	\centering
	\setlength{\tabcolsep}{2pt}
	\resizebox{1\linewidth}{!}{%
		\begin{tabular}{c|c|cc cccc}
			Methods& Train & CS &MAP & ADE-20&PC &PAS&PAS$^b$\\\toprule\midrule
			CAT-Seg*& CS&{\color{gray} 67.3} &26.7&26.6&45.7&73.6&65.7\\
			Decoder& CS& {\color{gray} 72.1}&26.8&27.4&48.1&78.4&70.0\\
			\rowcolor{Gray}
			FROVSS& CS& {\color{gray} 73.5}&28.8&28.8&48.6&78.9&70.6\\\midrule
			CAT-Seg*&MAP &61.5&{\color{gray} 52.6}&27.6&50.0&80.1&70.1\\
			Decoder& MAP&62.1&{\color{gray} 53.4}&28.4&50.4&85.5&73.9 \\
			\rowcolor{Gray}
			FROVSS& MAP&62.9&{\color{gray} 53.6}&28.9&51.3&85.5&74.0 \\\midrule
			CAT-Seg& ADE-20&41.2*&20.6*&{\color{gray} 46.8}& 46.7&85.5&70.3\\
			Decoder& ADE-20&42.8&22.3&{\color{gray}53.4}&56.6&93.2&75.7 \\
			\rowcolor{Gray}
			FROVSS& ADE-20&43.0&23.3&{\color{gray}53.6}&56.9&93.7&76.0 \\\midrule
			CAT-Seg& PC&35.8*&20.4*&23.0&{\color{gray} 62.4}&87.3&79.0\\
			Decoder& PC&36.1&21.9&29.3&{\color{gray} 67.4}&93.8&78.5\\
			\rowcolor{Gray}
			FROVSS& PC&37.4&22.4&29.3&{\color{gray} 68.3}&94.7&79.1\\\midrule
			CAT-Seg&COCO &41.7*&22.6*&27.2&57.5&93.7&78.3\\
			Decoder& COCO&44.0&23.1&28.6& 61.8&94.8&78.9\\
			\rowcolor{Gray}
			FROVSS& COCO&44.3&24.0&31.3& 67.4&95.6&79.8\\
			\bottomrule
	\end{tabular}}
	\caption{\textbf{OVSS performance comparison on different settings.}
		``Decoder'' stands for our proposed decoder results and FROVSS stands for our decoder with the proposed prompt augmentation and finetuning strategies. Our models demonstrate remarkable generalization capabilities even on visually different datasets. The scores evaluated on the same dataset used for training are colored in {\color{gray} gray} for clarity. Not reported results are trained by us and indicated by: ``*''.}
\label{tab:architectural_res_complete}
\vspace{-5mm}
\end{table}

\begin{table}[]
\centering
\setlength{\tabcolsep}{4pt}
\resizebox{1\linewidth}{!}{%
\begin{tabular}{c c c c c c c c c c}
Test & Train & CS & MAP& ADE-20& PC&Pas-20& PAS-20$^b$ \\\toprule\midrule
\rowcolor{Gray}
\multicolumn{8}{c}{Trained on Cityscapes}\\
& & 72.1&26.8&27.4&48.1&78.4&70.0\\
\checkmark& &72.0&27.2&27.6&48.3&78.9&70.3\\
\checkmark & \checkmark&72.3&28.1&28.8&48.6&78.9&70.6\\\midrule
\rowcolor{Gray}
\multicolumn{8}{c}{Trained on ADE-20}\\
& & 42.8&22.3&53.4&56.6&93.2&75.7 \\
\ \checkmark& &40.3&22.9&53.0&56.8&93.2&75.9\\
\checkmark & \checkmark&43.0&23.3&53.6&56.9&93.7&76.0\\\bottomrule
\end{tabular}}
\caption{\textbf{Performance comparison of prompt augmentation within the proposed FROVSS method.} Prompt augmentation is applied during only testing, training and testing or neither.
}
\label{tab:prompt}
\vspace{-5mm}
\end{table}

\paragraph{Prompt Augmentation}
Table \ref{tab:prompt} showcases the effectiveness of prompt augmentation described in Section \ref{sec:Prompt}. This technique can be applied during either the testing phase, the training phase, or both. Our findings indicate that utilizing prompt augmentation exclusively during testing enhances the model's generality at the expense of reduced specificity in relation to the training dataset. This is attributed to the decoder's increased specialization with the training-specific prompts. Conversely, implementing prompt augmentation throughout both training and testing phases enhances the model's performance in terms of both specificity and generality.

\begin{table}[]
	\centering
	\setlength{\tabcolsep}{3pt}
	\resizebox{1\linewidth}{!}{%
		\begin{tabular}{c c c| c c c c c c}
			\multicolumn{3}{c}{Prompt Aug}& \multicolumn{6}{c}{Evaluation dataset}\\
			Obj& PH & BG &CS& MAP& ADE-20& PC&Pas-20& PAS-20$^b$ \\\toprule
			\rowcolor{Gray}
			\multicolumn{9}{c}{Trained on: Cityscapes}
			\\
			&&&72.1& 26.8& 27.4& 48.1&78.4&70.0\\
			\checkmark&&&72.1& 27.2&27.6& 47.9&78.9& 70.3\\
			&\checkmark&&72.1& 25.7&27.4& 48.1&74.4& 69.0\\
			&&\checkmark&72.2& 27.6&27.7& 47.6&78.4& 70.0\\
			\checkmark&&\checkmark&72.3& 28.1&28.8& 48.6&78.9& 70.6
			\\\midrule
			\rowcolor{Gray}
			\multicolumn{9}{c}{Trained on: ADE-20}
			\\
			&&&42.8&22.3&53.4&56.6&93.2&75.7\\
			\checkmark&&&42.8&22.9&53.0&56.8&93.5&75.9\\
			&\checkmark&&42.7&22.3&53.4&56.6&93.2&75.7\\
			&&\checkmark&42.9&23.1&53.3&56.6&93.3&75.8\\
			\checkmark&&\checkmark&43.0& 23.3&53.6&56.9&93.7&76.0
			\\
			\bottomrule
		\end{tabular}}
		\caption{\textbf{Performance comparison of different prompt definition strategies.} Prompt augmentation is applied during training and testing. Key, Obj: Object, PH: Photometry, BG: Background.}
		\label{tab:performance_visual}
\end{table}

Moreover, Table \ref{tab:performance_visual} presents a comparison of performance across datasets using the proposed text augmentations studied. Notably, photometry of the images do not improve performance, indicating that further research is warranted, as such approaches have shown value in  image classification. We exclude results for photometry combinations with alternative augmentations, as they also failed to enhance performance. 

\begin{figure}[]
	\centering
	\begin{subfigure}[b]{0.45\linewidth}
		\includegraphics[clip, trim= 300 300 300 300, width=\linewidth]{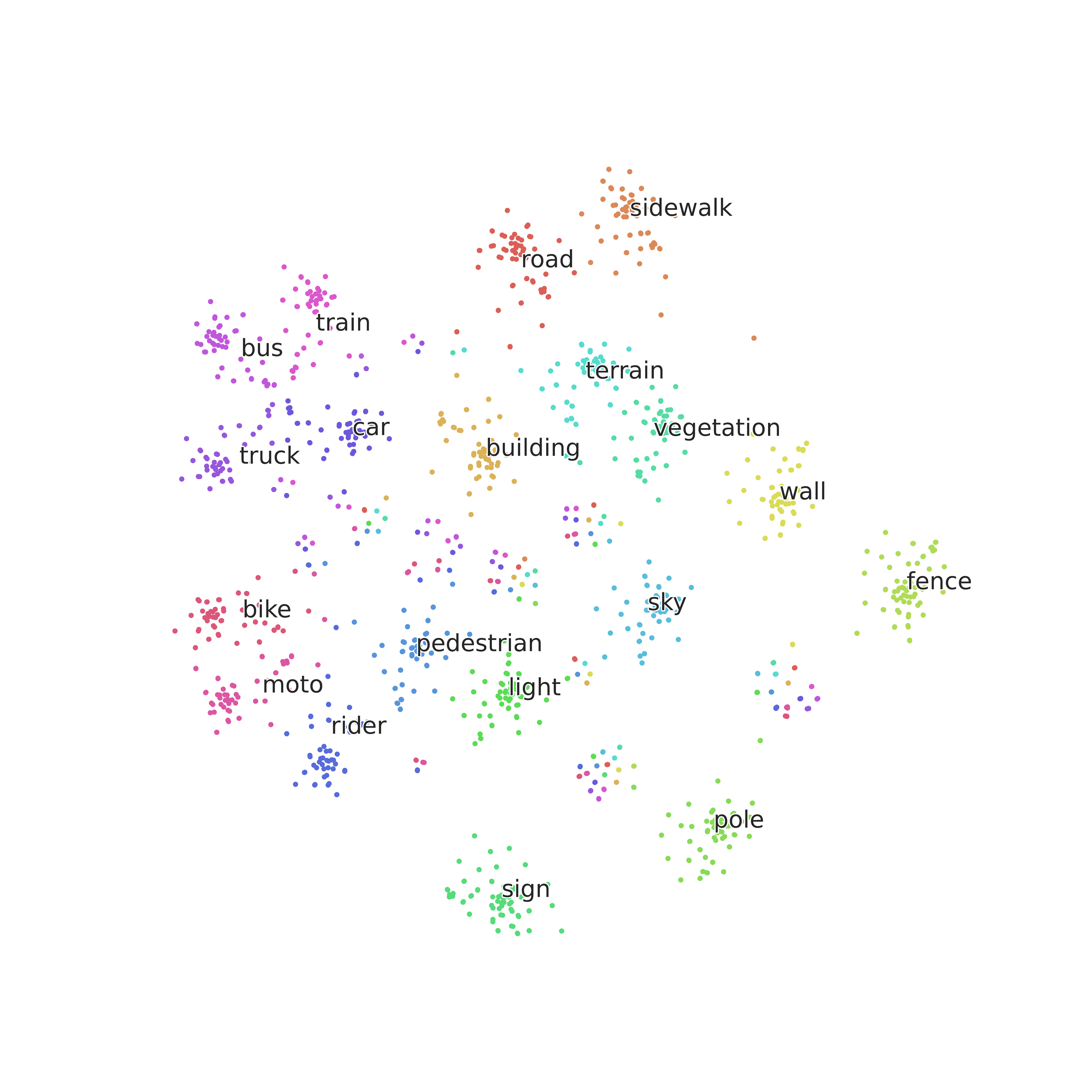}
		\caption{w/o text ensemble.}
	\end{subfigure}\hfill
	\begin{subfigure}[b]{0.45\linewidth}
		\includegraphics[clip, trim= 300 300 300 300, width=\linewidth]{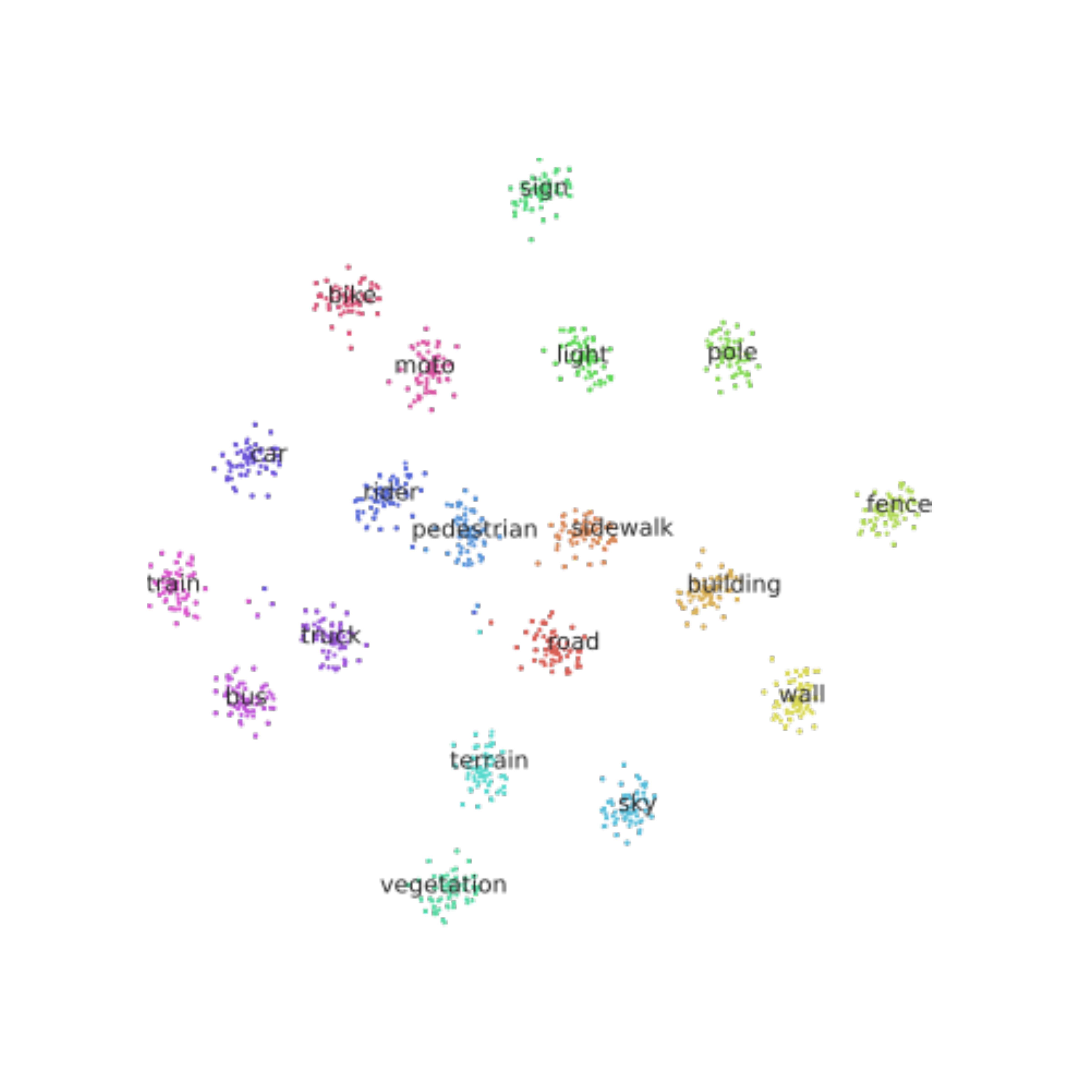}
		\caption{Text ensemble.}
	\end{subfigure}
	\caption{\textbf{Visual comparison of the text embeddings employed.} TSNE representation of the text embeddings employed to describe the 19 Cityscapes semantic categories.}
	\label{fig:TextualGuidance}
\end{figure}

\begin{figure}[]
		\centering
		\begin{subfigure}[b]{0.33\linewidth}
			\includegraphics[width=\linewidth]{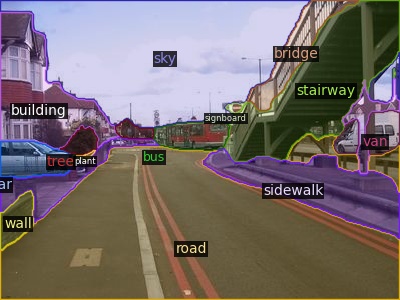}
		\end{subfigure}\hfill
		\begin{subfigure}[b]{0.33\linewidth}
			\includegraphics[width=\linewidth]{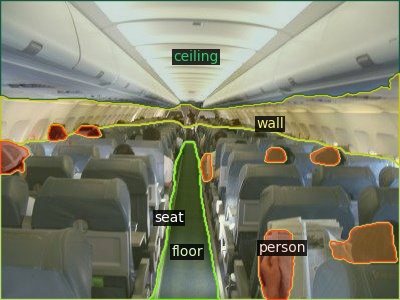}
		\end{subfigure}\hfill
		\begin{subfigure}[b]{0.33\linewidth}
			\includegraphics[width=\linewidth]{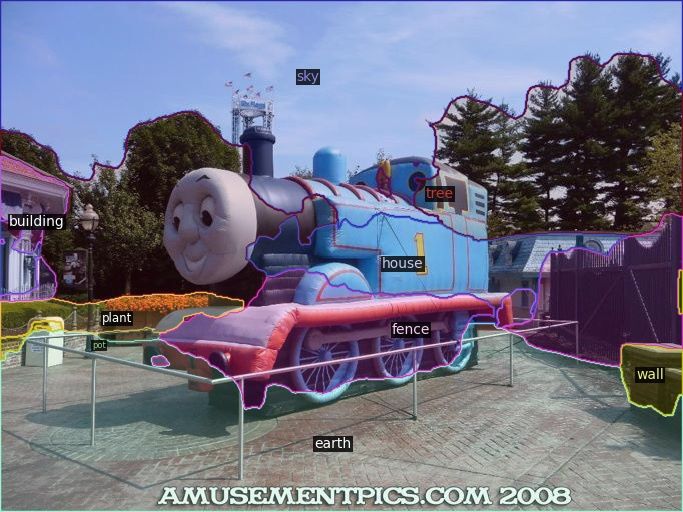}
		\end{subfigure}\hfill\\
		\begin{subfigure}[b]{0.33\linewidth}
			\includegraphics[width=\linewidth]{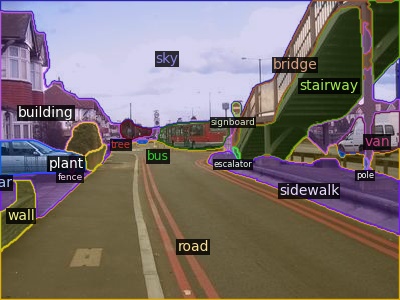}
		\end{subfigure}\hfill
		\begin{subfigure}[b]{0.33\linewidth}
			\includegraphics[width=\linewidth]{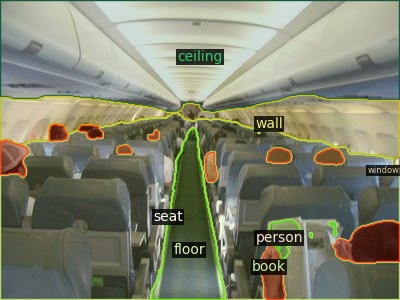}
		\end{subfigure}\hfill		
		\begin{subfigure}[b]{0.33\linewidth}
			\includegraphics[width=\linewidth]{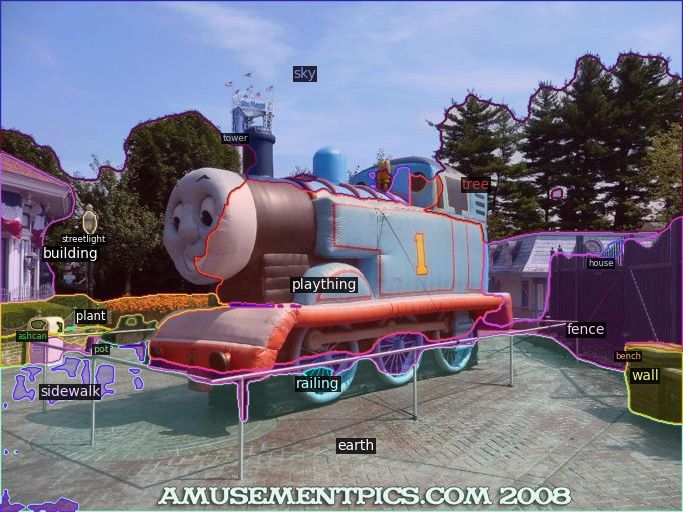}
		\end{subfigure}\hfill
		\caption{\textbf{Qualitative comparison of model trained and evaluated with (first row) and without (second row) prompt augmentation on the ADE-20 dataset.} The enhanced robust text embeddings allow the model to correctly segment the \textit{fence, wall} and \textit{plant} (first column), \textit{book} (second column) and \textit{plaything} (third column).}
		\label{fig:ADE-20}
	\end{figure}

Figure \ref{fig:TextualGuidance} compares the TSNE representation of the textual features employed. Notably, our robust text embeddings result in distinctly separated clusters for each class while maintaining logical inter-class relationships. For instance, while \textit{rider} and \textit{pedestrian} categories are closely grouped, \textit{rider} also aligns closely with \textit{bike} and \textit{moto}, whereas \textit{pedestrian} is positioned nearer to \textit{sidewalk}.  Moreover,  Figure \ref{fig:ADE-20},  illustrates qualitatively scenarios where the prompt augmentation allows the model to discern the true category; The \textit{fence, book} and the \textit{plaything} on each of the three columns column, which the model trained without the proposed textual features fails to identify them in favor of more common categories.

\paragraph{Finetuning of CLIP}
Table \ref{tab:finetune} compares different training protocols across three datasets suggesting that our fine-tuning enhances performance while preserving efficiency. We argue that applying a lower learning rate to the model's early layers yields improves performance by avoiding drastic changes that lead to misalignment's with the frozen text encoder. On the other hand, by selectively targeting at spatial relationship-targeted layers we reduce computational costs. 

\begin{table}[]
\centering
\setlength{\tabcolsep}{5pt}
\resizebox{1\linewidth}{!}{%
\begin{tabular}{l c c c c}
Fine-tuning & \# Parameters & CS  & ADE-20& PC\\\toprule\midrule
Full model&0.3B &69.9&50.2&61.0\\
Only decoder &2.3M&70.2&51.8&63.1\\
Spatial\cite{cho2023catseg} & 30M&72.1&53.6&67.4 \\
\rowcolor{Gray}
Proposed& 30M& 73.5& 54.1 &68.3\\\bottomrule
\end{tabular}}
\caption{\textbf{Ablation study on the fine-tuning strategy.} Reported results for the training dataset validation set. \# Parameters stands for the number of parameters tuned during training.} 
\label{tab:finetune}
\end{table}
\paragraph{Quantitative Comparison with State-Of-The-Art CLIP-based Segmentators}
To validate the first core contribution of our paper (FROVSS), we present Table \ref{tab:SOTA_CLIP} where the performance on different datasets of several CLIP-based segmentation methods is compared.  Additionally, in Table \ref{tab:SOTA_CLIP_OV} we compare the performance of our framework by comparing with other OVSS frameworks all trained on the COCO-stuff dataset. Please note that our model outperforms all other proposals, even the ones that employ additional foundational methods like \cite{Xu2023side} which employs SAM \cite{Kirillov_2023_ICCV}. 

\begin{table}[t]
\centering
\setlength{\tabcolsep}{3pt}
\resizebox{1\linewidth}{!}{%
\begin{tabular}{ll |c| c c c c}
Method &Venue&OV& CS  & ADE-20&PC & COCO \\\toprule\midrule
FreeSeg \cite{qin2023freeseg}&CVPR23& \checkmark&- &46.1&-&42.9\\
ECLIP \cite{li2022exploring}&ICLR23&\checkmark&- &-&-&25.8\\
CLIP sgr \cite{li2023clip}&arxiv23&\checkmark&31.4 &30.4&29.3&35.2\\
MasQCLIP \cite{xu2023masqclip}&ICCV23&\checkmark&- &30.4&57.8&47.3\\
\midrule
CLOUDS \cite{benigmim2023collaborating}&arXiv23&$\times$&60.2 &-&-&-\\
ZegCLIP \cite{Zhou_2023_CVPR}&CVPR23& $\times$&- &-&46.5&40.7\\\midrule\midrule
\rowcolor{Gray}
FROVSS& &\checkmark & \textbf{73.5}&  \textbf{53.6}&\textbf{68.1}&\textbf{48.8}\\\bottomrule
\end{tabular}}
\caption{\textbf{CLIP-based segmentators performance comparison in supervised setting}. Our models outperforms by large margins CLIP-based segmentators across five different datasets. Each model is trained and evaluated on the dataset indicated.  Not reported results are indicated by: ``-''. }
\label{tab:SOTA_CLIP}
\vspace{-5mm}
\end{table}

\begin{table}[]
\centering
\setlength{\tabcolsep}{5pt}
\resizebox{1\linewidth}{!}{%
\begin{tabular}{l l  c c c}
Method& Venue &ADE-20& PC& PAS-20 \\\toprule\midrule
FreeSeg  \cite{qin2023freeseg}&CVPR23&24.6&-&91.9\\
SegCLIP  \cite{Luo2023SegCLIP}&ICML23&-& 24.7&52.6\\
OVSeg \cite{liang2023open}&CVPR23 &24.8&53.3&92.6\\
SAN  \cite{Xu2023side}&CVPR23&27.5&53.8&94.0\\
HIPIE  \cite{hipie}&NeurIPS23&29.0&-&63.8\\
CAT-seg \cite{cho2023catseg}&CVPR24&27.2 &57.5&93.7\\\midrule\midrule
\rowcolor{Gray}
FROVSS&& \textbf{31.3}&\textbf{67.4}&\textbf{95.6}\\\bottomrule
\end{tabular}}
\caption{\textbf{Performance comparison of open vocabulary semantic segmentation frameworks}. Our method significantly outperforms current alternatives by 8\% mIoU, 17\% mIoU and 2\% on the ADE-20, Pascal-Context and Pascal VOC datasets respectively. Not reported results are indicated by: ``-''.}
\label{tab:SOTA_CLIP_OV}
\vspace{-7mm}
\end{table}

\paragraph{Qualitative Comparison with State-Of-The-Art CLIP-based Segmentators}
\begin{figure}[h]
	\centering
	\begin{subfigure}[b]{.3\linewidth}
		\caption{Color Image}
		\includegraphics[clip, trim= 0 1300 2300 600, width=\linewidth]{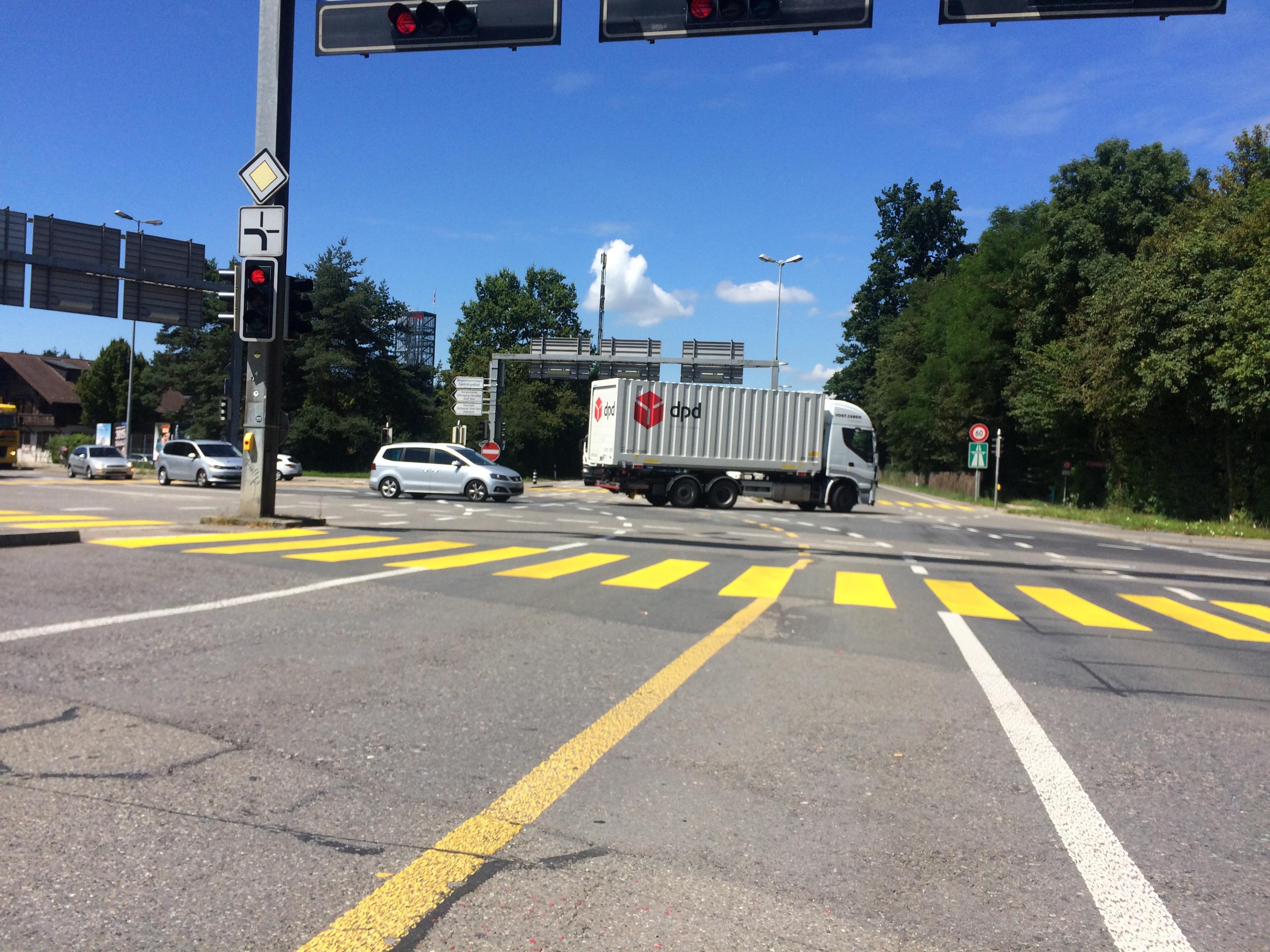}
	\end{subfigure}
	\begin{subfigure}[b]{.3\linewidth}
		\caption{CAT-Seg \cite{cho2023catseg}}
		\includegraphics[clip, trim= 0 970 1720 450, width=\linewidth]{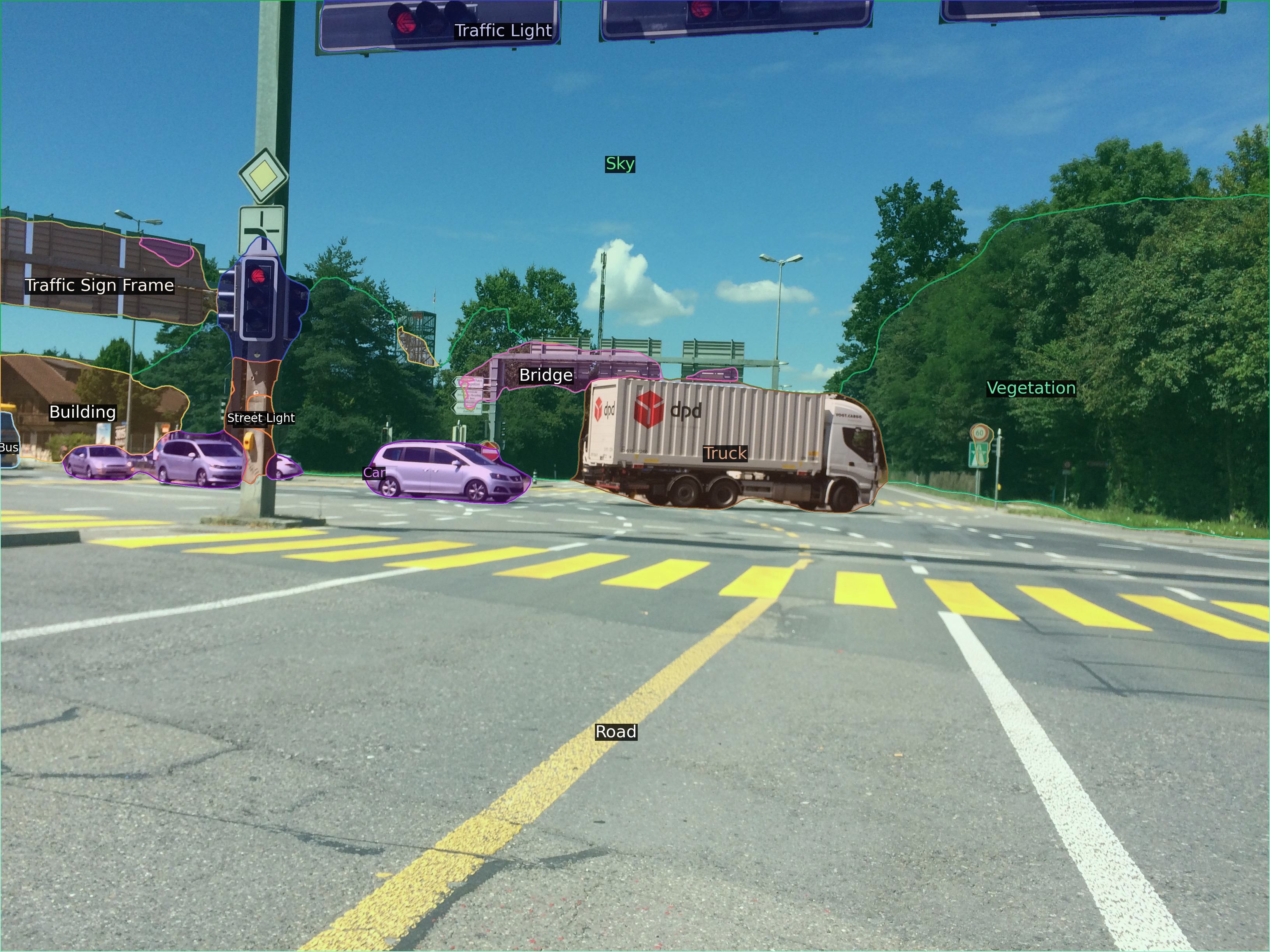}
	\end{subfigure}
	\begin{subfigure}[b]{.3\linewidth}
		\caption{Ours}
		\includegraphics[clip, trim= 0 970 1720 450, width=\linewidth]{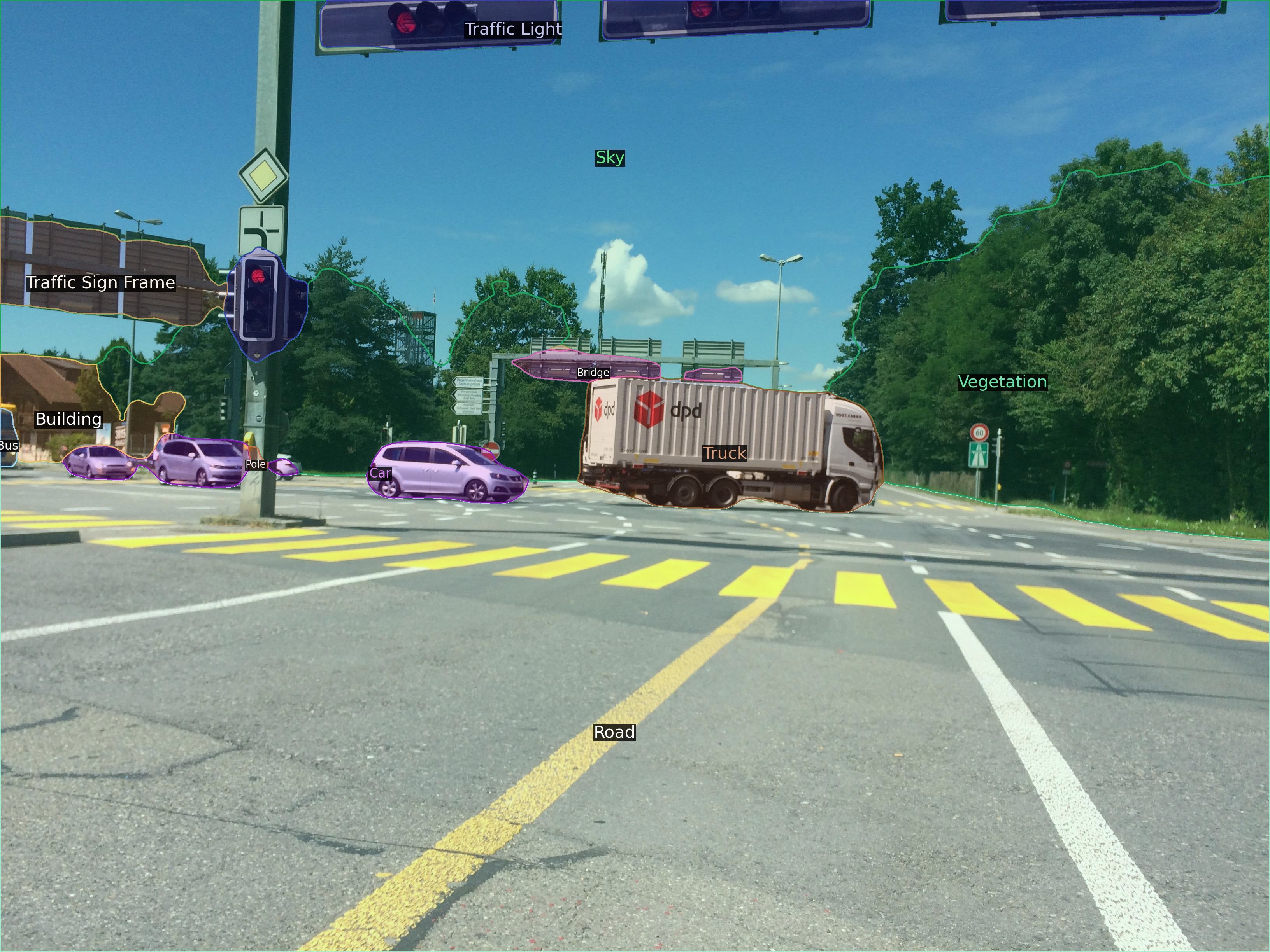}
	\end{subfigure}\\
	\begin{subfigure}[b]{.3\linewidth}
		\includegraphics[clip, trim= 90 150 100 100, width=\linewidth]{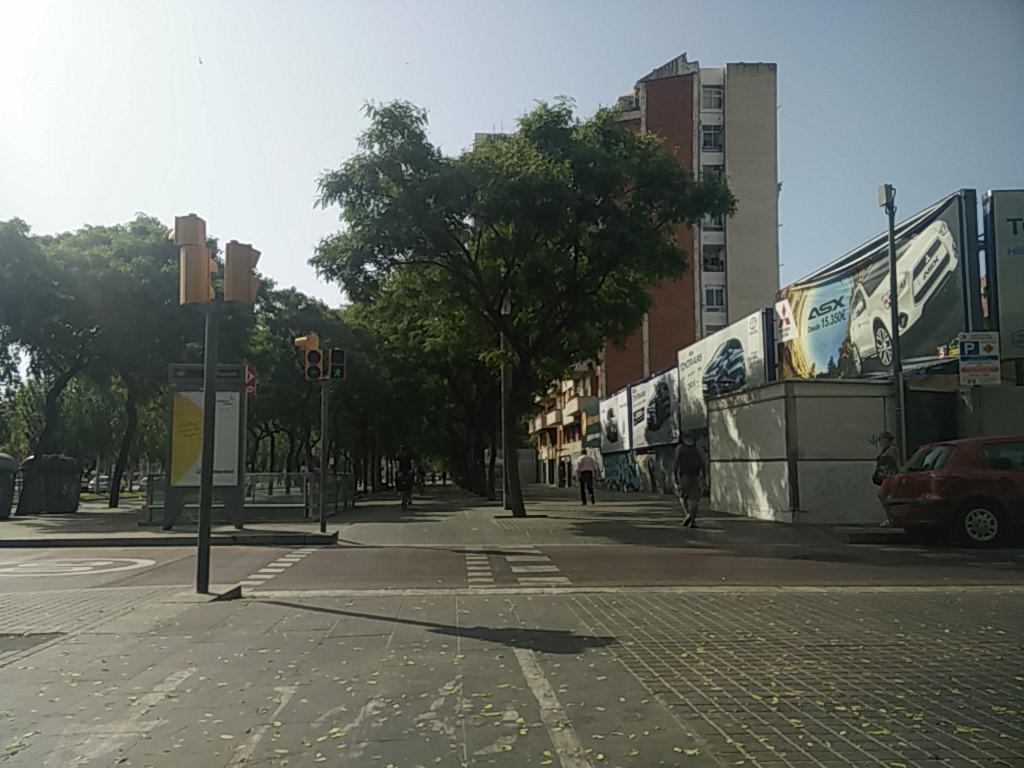}
	\end{subfigure}
	\begin{subfigure}[b]{.3\linewidth}
		\includegraphics[clip, trim= 90 150 100 100, width=\linewidth]{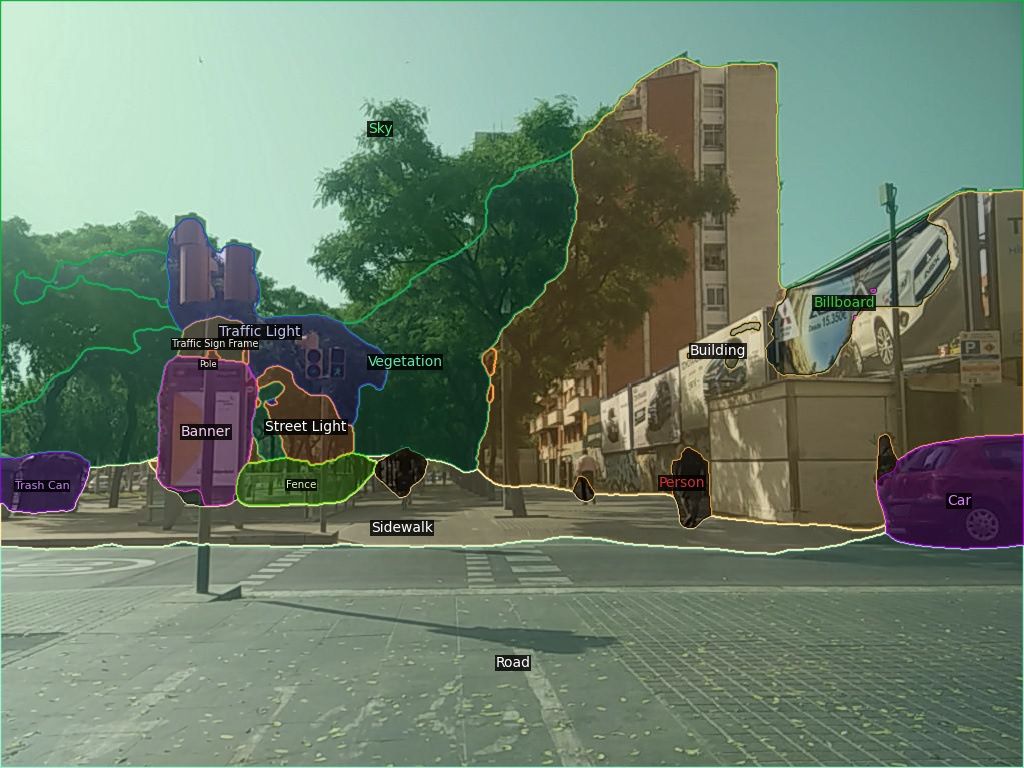}
	\end{subfigure}
	\begin{subfigure}[b]{.3\linewidth}
		\includegraphics[clip, trim= 90 150 100 100, width=\linewidth]{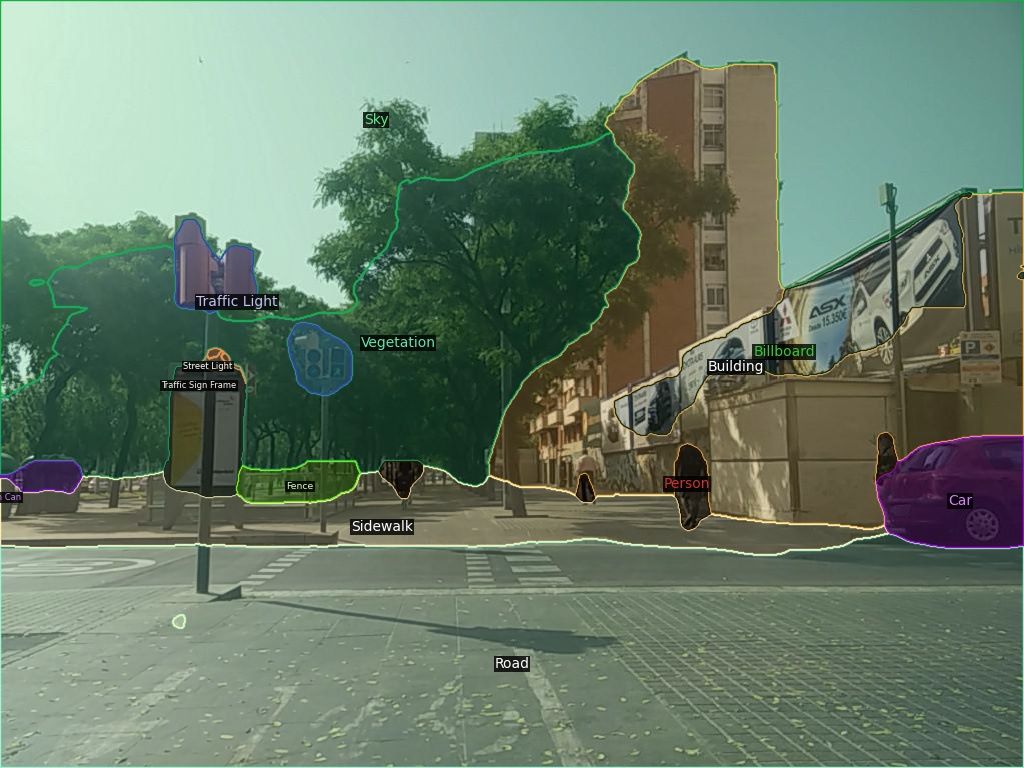}
	\end{subfigure}
	
	\caption{\textbf{Visual performance comparison of CAT-seg and our model on the Mapillary dataset, both trained only with COCO.} Our model presents better segmentation of the \textit{vegetation, building} and \textit{traffic light}.}
	\label{fig:OVSS_mapillary}
	\vspace{-5mm}
\end{figure}
Figure \ref{fig:OVSS_mapillary} presents a qualitative comparison of the state-of-the-art OVSS model CAT-seg \cite{cho2023catseg} on the Mapillary dataset, both trained only on the COCO dataset. We have selected this dataset as Mapillary is a densely segmented dataset across multiple labels. As our ASPP and prompt augmentation techniques are tailored to improve on densely labeled areas and recognition of visually similar categories, Mapillary stands as the best testing ground. Notably, our model outperforms CAT-seg both in recognition and level of segmentation detail. Specifically,  in the first row results we find our model presents better segmentation of the \textit{vegetation, building} and \textit{frame}. Additionally, our model does not misclassify the \textit{traffic sign} facing backwards with a bridge. On the Second row, we find better segmentation for \textit{vegetation, building} and \textit{sky}. Similarly, the center of the image is densely populated and our model is capable of accurately segment each class meanwhile, CAT-seg fails to segment the \textit{traffic sign, lights} and \textit{frame}.

\begin{figure}[]
	\centering
	\begin{subfigure}[b]{.23\linewidth}
		
		\includegraphics[clip, trim= 0 53 150 13, width=\linewidth]{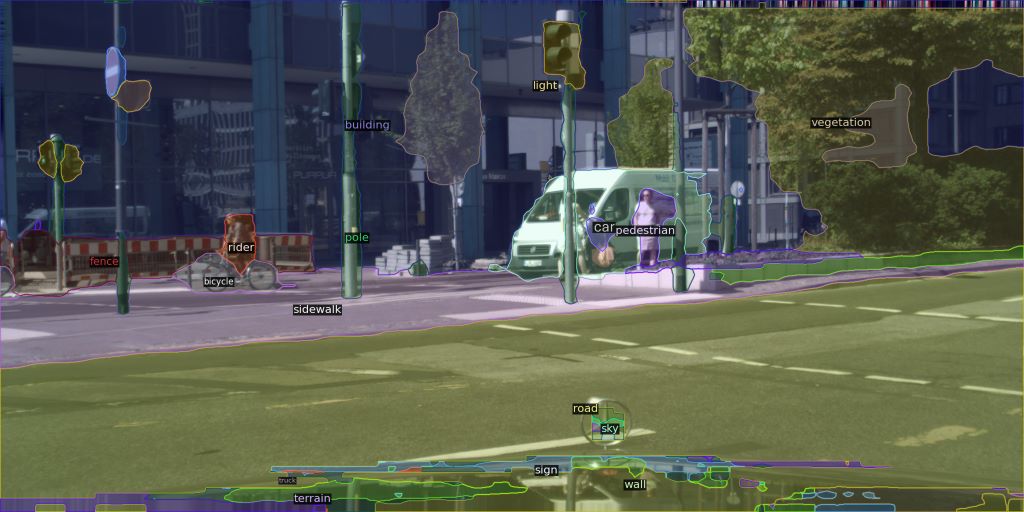}
	\end{subfigure}\hfill
	\begin{subfigure}[b]{.38\linewidth}
		\includegraphics[clip, trim= 120 42 24 40, width=\linewidth]{images/Supplementary/Comparative/Cityscapes/frankfurt_000001_007407_leftImg8bitbase.jpg}
	\end{subfigure}\hfill
	\begin{subfigure}[b]{.38\linewidth}
		\includegraphics[clip, trim= 13 50 112 25, width=\linewidth]{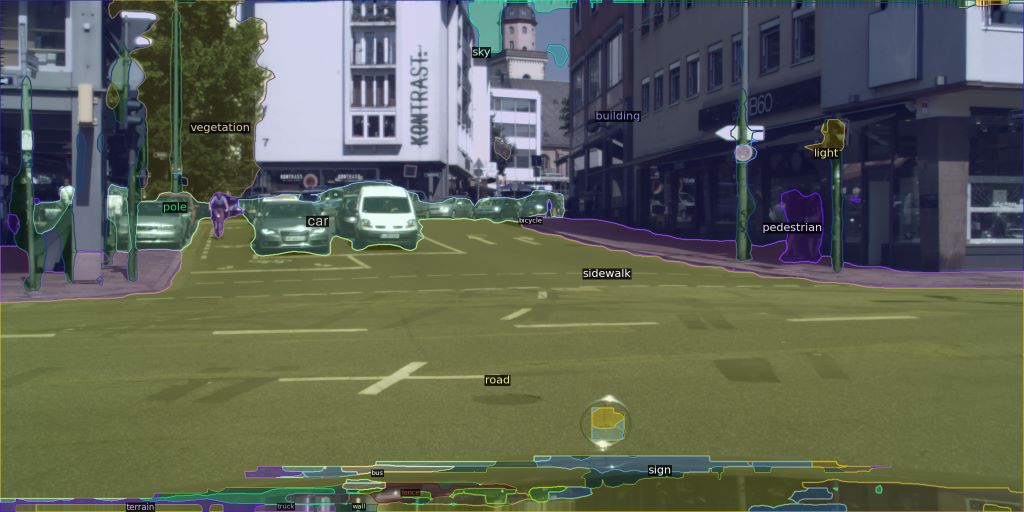}
	\end{subfigure}\hfill
	\\
	\begin{subfigure}[b]{.23\linewidth}
		\includegraphics[clip, trim= 0 53 150 13, width=\linewidth]{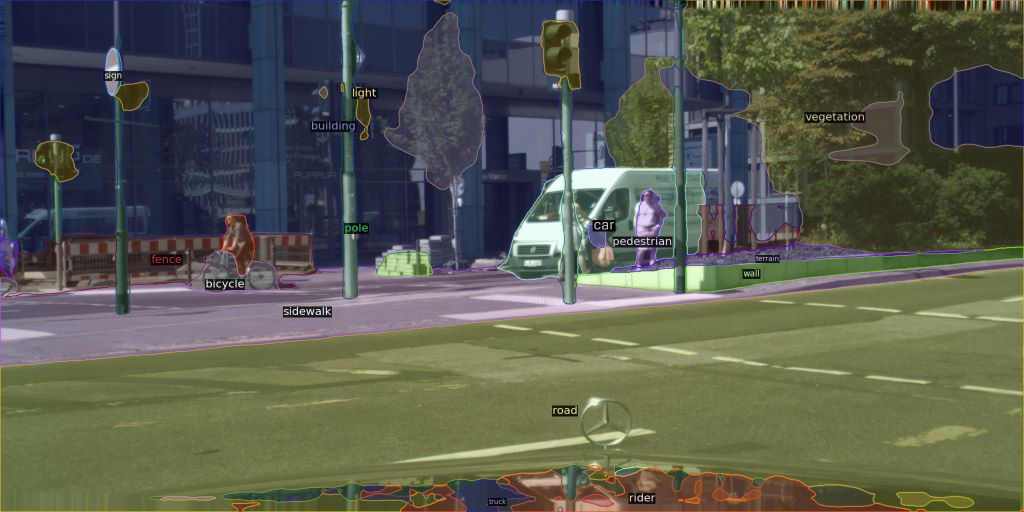}
	\end{subfigure}\hfill
	\begin{subfigure}[b]{.38\linewidth}
		\includegraphics[clip, trim= 120 42 20 40, width=\linewidth]{images/Supplementary/Comparative/Cityscapes/frankfurt_000001_007407_leftImg8bit.jpg}
	\end{subfigure}\hfill
	\begin{subfigure}[b]{.38\linewidth}
		\includegraphics[clip, trim= 13 50 110 25, width=\linewidth]{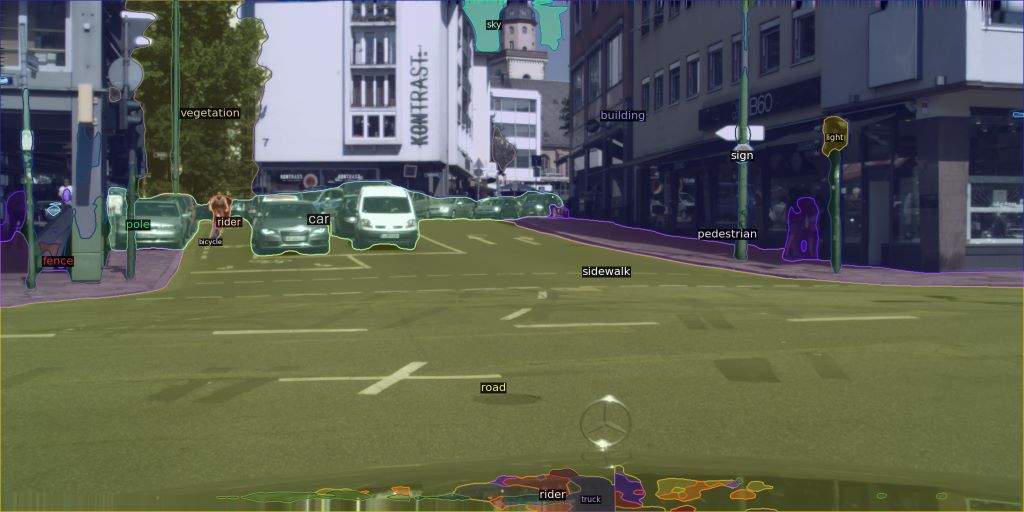}
	\end{subfigure}
	\caption{\textbf{Visual performance comparison of CAT-seg (first row) and our model (second row) both trained and evaluated only with the Cityscapes dataset.} We find that our model presents better segmentation of \textit{rider}, \textit{bicycle}, \textit{fence}, \textit{light} (first column), \textit{wall},  \textit{pole},  \textit{pedestrian} (second column). Additionally, our model does not misclassify the \textit{sign} (first column) and \textit{rider} and \textit{bike} with \textit{sidewalk} (third column).}
	\label{fig:OVSS_City}
\end{figure}

\begin{figure}[]
	\centering
	\begin{subfigure}[b]{.35\linewidth}
		\includegraphics[width=\linewidth]{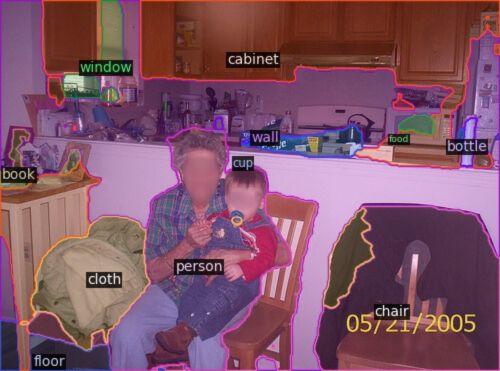}
	\end{subfigure}\hfill
	\begin{subfigure}[b]{.3\linewidth}
		\includegraphics[clip, trim= 0 70 0 100, width=\linewidth]{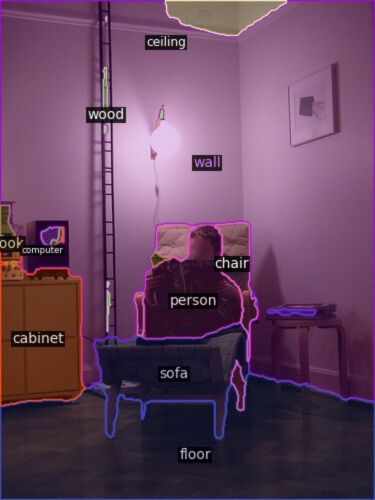}
	\end{subfigure}\hfill
	\begin{subfigure}[b]{.34\linewidth}
		\includegraphics[clip, trim= 0 30 0 150, width=\linewidth]{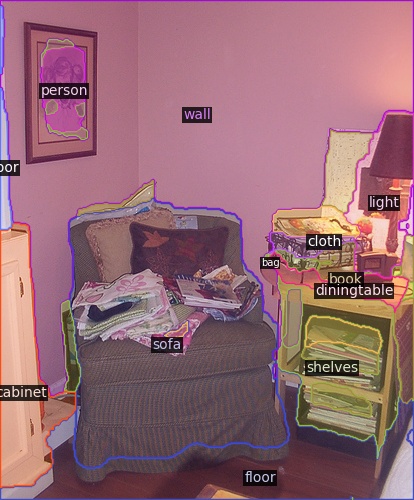}
	\end{subfigure}\hfill

	\begin{subfigure}[b]{.35\linewidth}
		\includegraphics[width=\linewidth]{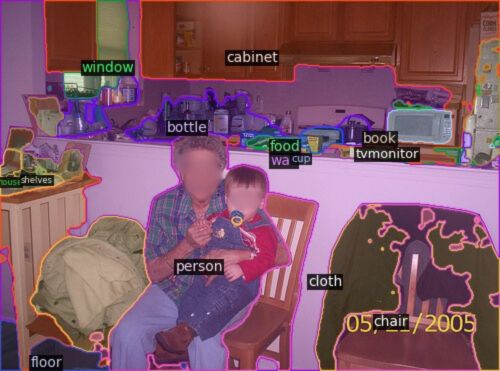}
	\end{subfigure}\hfill
	\begin{subfigure}[b]{.3\linewidth}
		\includegraphics[clip, trim= 0 70 0 100, width=\linewidth]{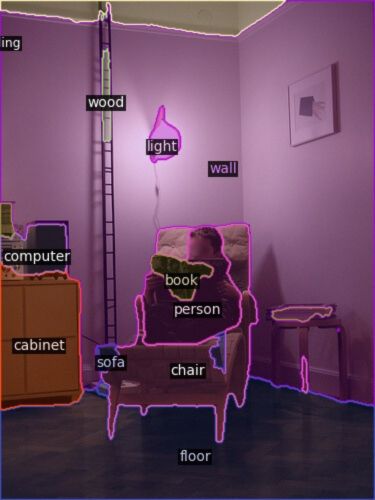}
	\end{subfigure}\hfill
	\begin{subfigure}[b]{.34\linewidth}
		\includegraphics[clip, trim= 0 30 0 150, width=\linewidth]{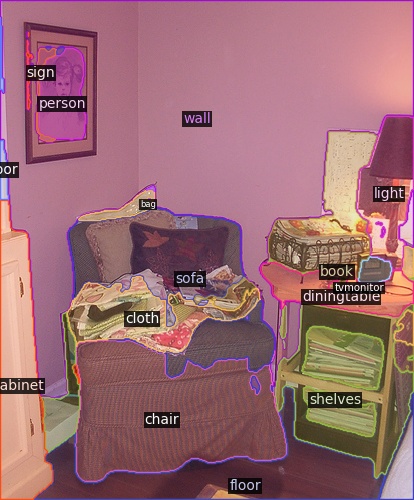}
	\end{subfigure}
	\caption{\textbf{Visual performance comparison of CAT-seg (first row) and our model (second row) on the Pascal Context validation set, both trained only with the Pascal Context training set.} Notably our model does not only find more instances but also is less prone to misclassify categories. Please note that people's faces have been intentionally blurred to preserve anonimity.}
	\label{fig:OVSS_Pascal}
\end{figure}

To further validate our models improved segmentation of fine details, we present a comparison in Figure \ref{fig:OVSS_City} on the Cityscapes validation set both trained only with the training set of the Cityscapes dataset. Moreover, to validate the improved recognition of our model we showcase a comparison on the Pascal Context validation set in Figure \ref{fig:OVSS_Pascal} both models trained only with the training set of the Pascal Context dataset.

\paragraph{Visual Guidance Encoder Analysis}
Table \ref{tab:VisualGuidance} compiles the performances resulting from employing different auxiliary image encoders. We find that even untrained features help in the segmentation as randomly initialized auxiliary encoders outperform not employing guidance. However, it seems that the choice of randomly initialized encoder has little impact. On the other hand, as expected, pre-trained visual encoders yield significantly better results. Moreover, similarity with the pre-training task is also an important factor for performance, obtaining the best results with the SAM \cite{Kirillov_2023_ICCV} backbone. Note that our reported results in the state-of-the-art comparison do not employ SAM guidance for fairness. 

\begin{table}[]
\centering
\setlength{\tabcolsep}{2pt}
\resizebox{1\linewidth}{!}{%
\begin{tabular}{l c c c c c c}
Encoder & Objective & ADE-20& CS & MAP&  PC&PAS$^b$   \\\toprule\midrule
No guidance&-& 49.1& 36.0&16.3 & 42.1& 73.7\\
ViT-L \cite{dosovitskiy2021an}& - & 50.0& 38.2&20.1 & 44.9&74.6 \\
Swin \cite{liu2021swinv2}& - & 49.3& 38.1&19.9 & 44.5&73.8 \\
ViT-L\cite{dosovitskiy2021an} & SAM \cite{Kirillov_2023_ICCV}& 54.9 & 50.3& 29.4& 60.1& 80.0 \\
\rowcolor{Gray}
Swin \cite{liu2021swinv2}& Classification & 53.6&43.0&23.3&56.9&76.0\\\bottomrule
\end{tabular}}
\caption{\textbf{Ablation study on the shape guidance image encoder for the ADE-20 dataset.} Randomly initialized encoders are denoted by ``-'' on the Objective column.}
\label{tab:VisualGuidance}
\end{table}

\subsection{Results on OVSS enhanced with UDA}
In this subsection, we explore including UDA techniques (see Section \ref{uda} and Figure \ref{fig:Framework}), leveraging unlabeled color images from a given target domain for training. 
\begin{table}[]
	\centering
	\setlength{\tabcolsep}{5pt}
	\resizebox{1\linewidth}{!}{%
		\begin{tabular}{c c c c c c }
			Teacher &  Mixup& ASPP & Prompt & Finetuning & mIoU  \\\toprule\midrule
			$\times$ & $\times$ & $\times$ & $\times$& $\times$ & 36.2\\
			\checkmark & $\times$ & $\times$ & $\times$ & $\times$ & 44.7\\
			\checkmark & \checkmark & $\times$ & $\times$ & $\times$ & 51.1\\
			\checkmark & \checkmark & \checkmark & $\times$ & $\times$ & 54.8\\ 
			\checkmark & \checkmark & \checkmark & \checkmark & $\times$& 60.0\\
			\rowcolor{Gray}
			\checkmark & \checkmark & \checkmark & \checkmark & \checkmark & 61.5 \\
			\bottomrule
	\end{tabular}}
	\caption{\textbf{Ablation study on the Synthia-to-Cityscapes setup.} Model trained with labeled Synthia images and unlabeled Cityscapes color images. mIoU evaluated on the Cityscapes validation set. Prompt stands for our proposed prompt augmentation (see Figure \ref{fig:prompts}) and Finetuning stands for our proposed finetuning (see Equation \ref{eq:finetuning})} 
	\label{tab:UDAablation}
\end{table}

\begin{table*}[]
	\centering
	\setlength{\tabcolsep}{2pt}
	\resizebox{\textwidth}{!}{%
		\begin{tabular}{l| c c c c c c c c c c c c c c c c c c c | c}
			Method &\rotatebox[origin=c]{90}{\textit{road} } & \rotatebox[origin=c]{90}{\textit{sidewalk} } & \rotatebox[origin=c]{90}{\textit{building} } & \rotatebox[origin=c]{90}{\textit{wall}} &  \rotatebox[origin=c]{90}{\textit{fence}}& \rotatebox[origin=c]{90}{\textit{pole} } &\rotatebox[origin=c]{90}{\textit{light}}&\rotatebox[origin=c]{90}{\textit{sign} } & \rotatebox[origin=c]{90}{\textit{vegetation}} & \rotatebox[origin=c]{90}{\textit{terrain}*}& \rotatebox[origin=c]{90}{\textit{sky}}& \rotatebox[origin=c]{90}{\textit{pedestrian}}& \rotatebox[origin=c]{90}{\textit{rider}} & \rotatebox[origin=c]{90}{\textit{car}}& \rotatebox[origin=c]{90}{\textit{truck}*} & \rotatebox[origin=c]{90}{\textit{bus}}& \rotatebox[origin=c]{90}{\textit{train}*} & \rotatebox[origin=c]{90}{\textit{motorcycle}} & \rotatebox[origin=c]{90}{\textit{bicycle}}& mean\\\midrule\midrule
			
			Source Only&88.1&54.1&82.9&19.8&6.9&26.0&13.3&21.7&82.0&0.0&87.1&48.9&15.4&68.0&0.4&36.8&0.0&12.5&30.1&36.2\\
			Full E.M.A.& 93.5&63.1&86.1&43.8&24.3&21.8&35.5&39.0&85.2&0.0&91.0&60.3&31.2&82.5&59.4&61.9&0.1&31.4&49.5& 50.5\\\rowcolor{Gray}
			UDA-FROVSS &96.1&68.7&88.8&52.4&36.5&28.9&47.6&46.5&87.9&0.0&93.0&70.1&38.9&89.1&80.3&80.2&60.2&45.6&58.8&61.5\\
			\bottomrule
	\end{tabular}}
	\caption{\textbf{Per-class performance on the Synthia to Cityscapes domain adptation setup.} Target private categories are indicated by ``*''. Notably, the full update of the teacher model leads to the forgetting of the target private category: \textit{train}. }
	\label{tab:UDA_Synthia}
\end{table*}
\paragraph{Ablation Study on the UDA Techniques.}
\begin{table}[]
	\centering
	\resizebox{1\linewidth}{!}{%
		\begin{tabular}{l| c c c c c }
			Source & \multicolumn{5}{c}{Target} \\
			&CS & MAP& ADE-20& PC& PAS$^b$\\\midrule\midrule
			CS&{\color{gray}73.5}&33.0&32.3&49.8&70.2\\
			MAP&72.6&{\color{gray}53.8}&31.7&54.4&74.1\\
			ADE-20&72.5&24.6&{\color{gray}54.1}&57.4&77.8\\
			PC&44.6&24.2&30.1&{\color{gray}68.3}&80.6\\
			PAS-20$^b$&18.0&7.6&19.5&39.5&{\color{gray}84.9}\\\bottomrule
		\end{tabular}}
	\caption{\textbf{Performance comparison of the UDA-FROVSS models.} Models trained on labeled source data and unlabeled target data, evaluated on the target validation set. Supervised performance is indicated in {\color{gray}gray} in the diagonal.}
	\label{tab:UDA}
\end{table}

Table \ref{tab:UDAablation} presents the ablation results for the different components presented in the paper. We notice that Prompt Augmentation and Image MixUp significantly drive performance. First, the definition of the prompts help the transfer of categories cross-domain. Second, Image MixUp acts as a data augmentation technique helping the model learn the semantic edges of objects and improving its classification accuracy. As a limitation, we find that our model is not able to distinguish between \textit{terrain} (not included in the Synthia dataset) and \textit{vegetation} (semantically related and highly prevalent in the Synthia dataset).

\subsection{Teacher Update}
\begin{figure}[]
	\centering
	\begin{subfigure}[b]{\linewidth}
		\centering
		
		\includegraphics[clip, trim= 0 400 500 100
		,width=\linewidth]{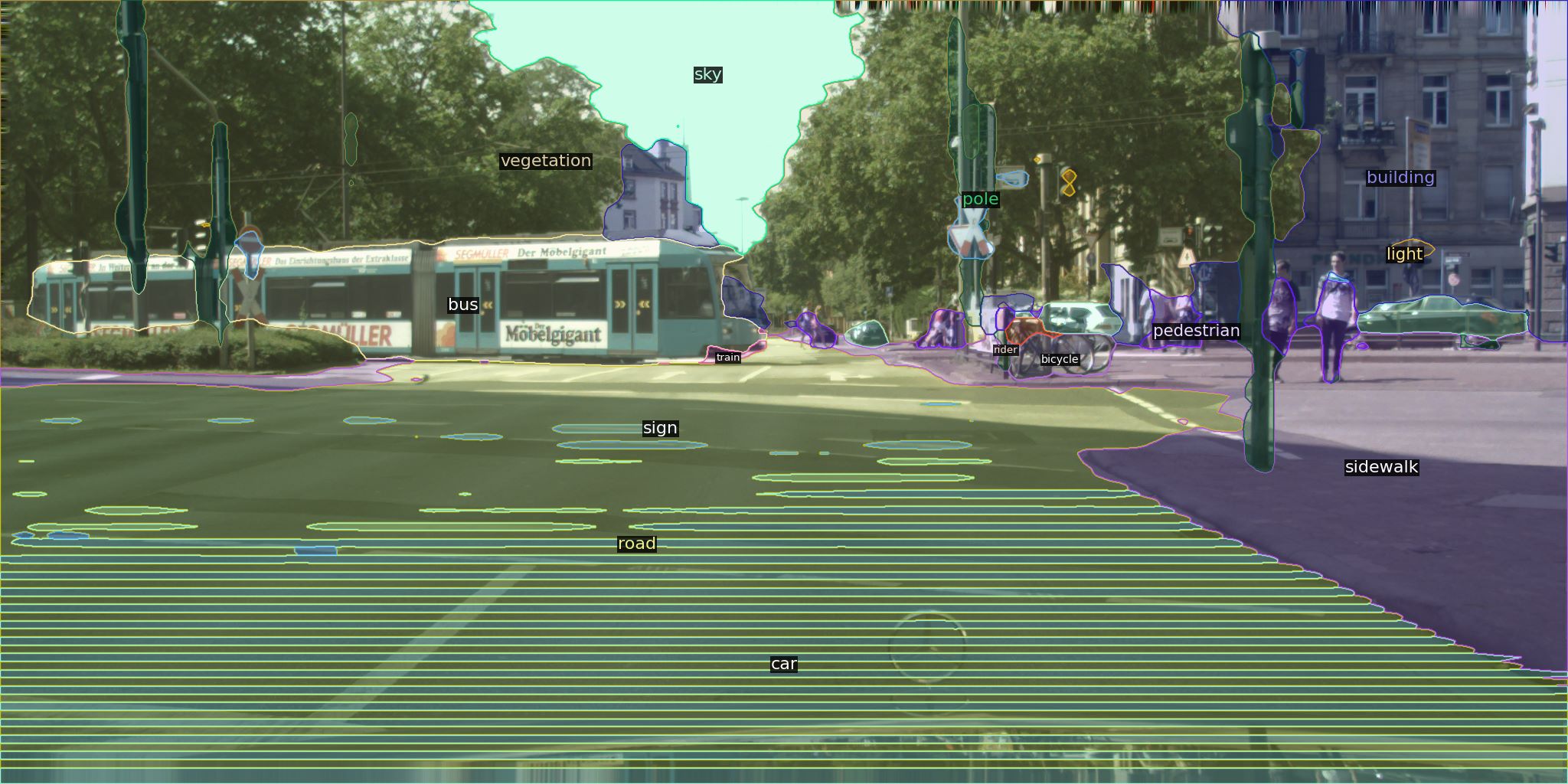}   
		\caption{Full E.M.A. update of the teacher}
	\end{subfigure}
	\\\vspace{2mm}
	\begin{subfigure}[b]{\linewidth}
		\centering
		
		\includegraphics[clip, trim= 0 400 500 100
		,width=\linewidth]{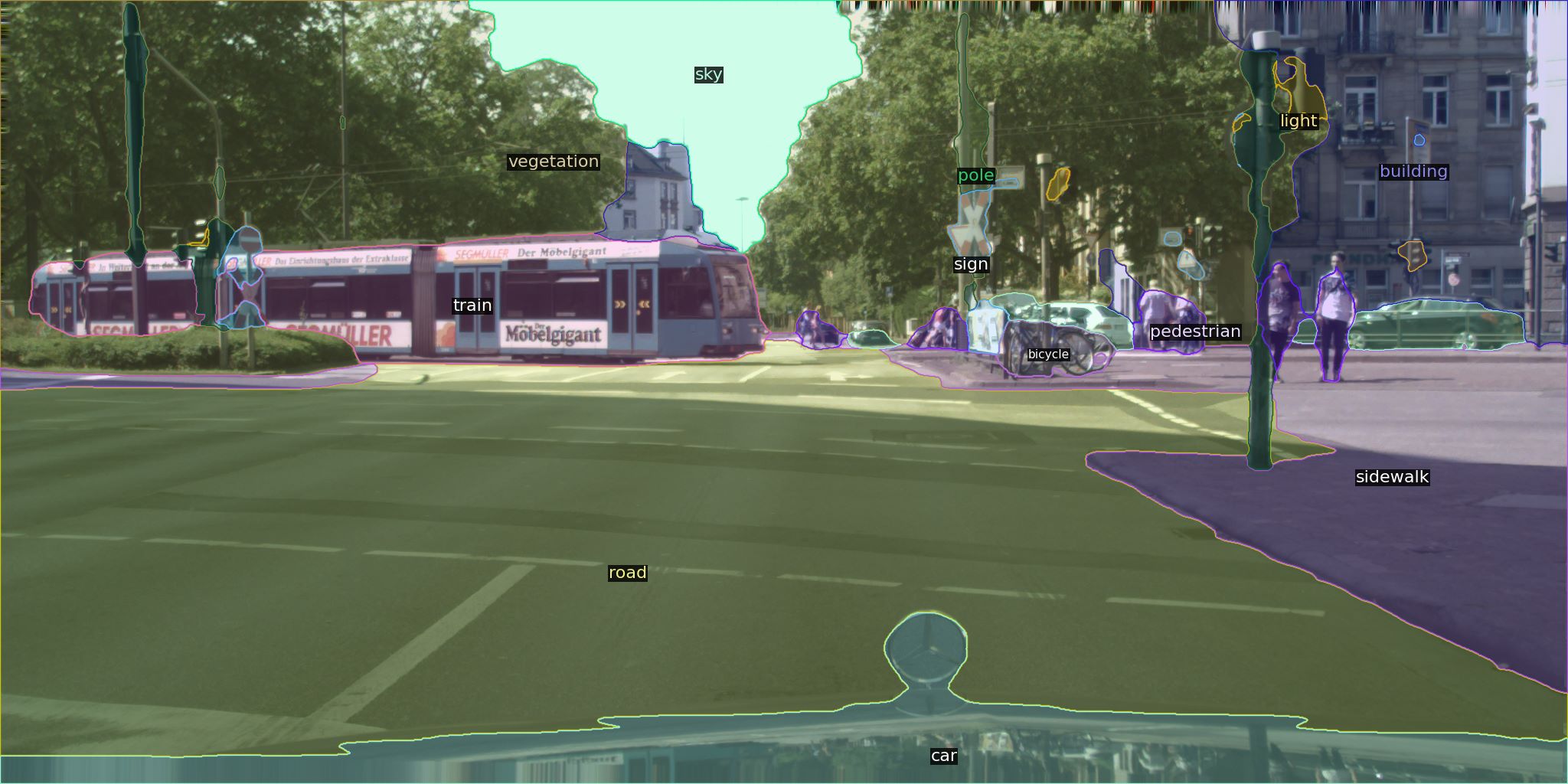}   
		\caption{Proposed Teacher-Student update}
	\end{subfigure}
	\caption{\textbf{Example of the full teacher update against our UDA-FROVSS proposed teacher update.} Our model remains capable of segmenting the target private \textit{train} category after training.}
	\label{fig:train}
\end{figure}

In our work we decided to only update the teacher decoder. This is mainly due to two reasons: First, CLIP has shown remarkable zero shot segmentation performance \cite{xu2023masqclip}. Second, we have found that early iterations with AdamW optimizer lead to the model over-fitting to the training dataset. In our first approximation, we changed the optimizer to SGD, which helped alleviating such over-fitting. However, this seemed to affect the encoded knowledge, leading to worse performing models.  Therefore, we opted to only update the decoder of the teacher while maintaining the encoder frozen. This helped the teacher model preserving the encoded knowledge of the target private labels thus allowing the student to learn from those target private categories. However, such update led to a misalignment between the teacher encoder and decoder, that we aim to explore in future work. Thereby, we opted to temporally reduce the impact of teacher predictions in the student training, by only taking into account teacher predictions for the initial iterations and progressively incorporating the student predictions in the pseudo-labels, hence defining the combination of teacher and student labels described in equation 6 of the paper. Table \ref{tab:UDA_Synthia} and Figure \ref{fig:train} motivate the employment of a specific teacher-student framework to preserve recognition of unseen categories.

\begin{figure}[]
	\centering
	\begin{subfigure}[b]{.48\linewidth}
		\includegraphics[width=\linewidth]{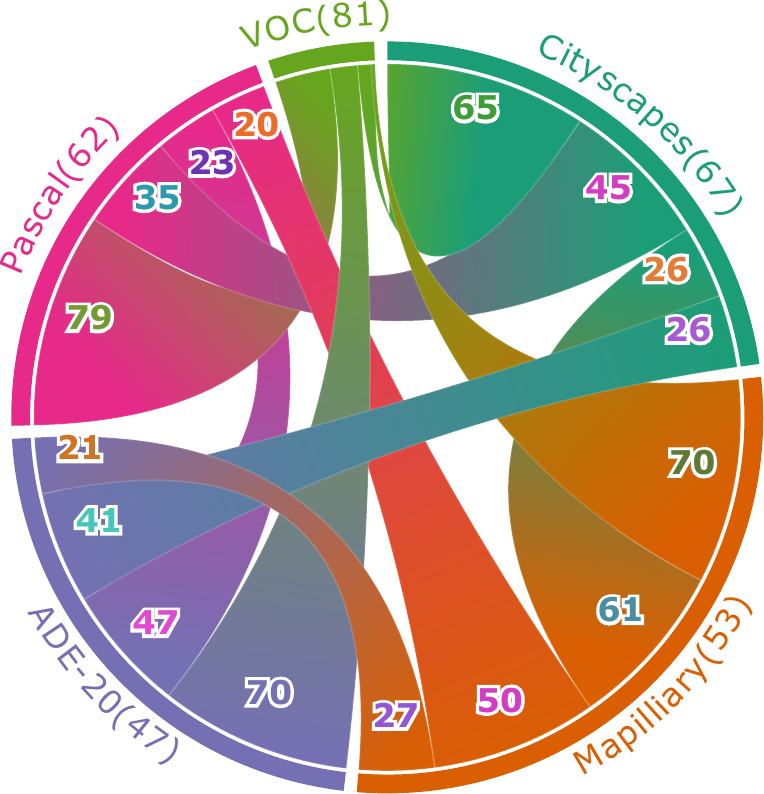}
		\caption{CAT-Seg \cite{cho2023catseg}}
	\end{subfigure}\hfill
	\begin{subfigure}[b]{.48\linewidth}
		\includegraphics[width=\linewidth]{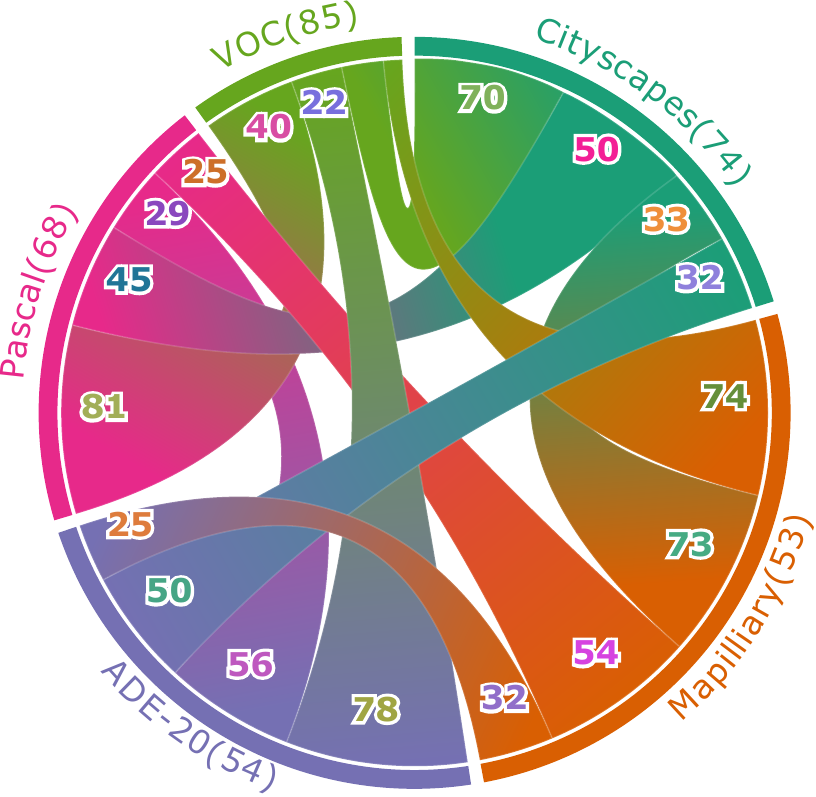}
		\caption{UDA-FROVSS}
	\end{subfigure}
	\caption{\textbf{Performance comparison across five datasets against the state-of-the-art CAT-Seg \cite{cho2023catseg} for Open Vocabulary Semantic Segmentation (OVSS)}. \textit{UDA-FROVSS} correspond to our proposal for UDA based on VLMs. The numbers adjacent to dataset names indicate performance when training and testing with the same dataset. The numbers inside indicate testing results when models are evaluated on datasets at the opposite ends of the chords. For instance, as seen in subfigure (b), our method achieves a 32 mIoU when trained on Cityscapes and tested on ADE-20. } 
	\label{figcircles}
\end{figure}

\paragraph{Performance Across Diferent Real Datasets.}
Table \ref{tab:UDA} presents the results of our UDA-enhanced models. As expected, performance improvements seem to be highly related to the target and source similarity. Notably, Cityscapes and Mapillary are closely related. Therefore, employing Cityscapes as the source domain presents the best performance on the Mapillary dataset. These results are summarized and compared with  CAT-seg \cite{cho2023catseg} in Figure \ref{figcircles}. Notably, CAT-seg models trained with a small dataset as VOC significantly underperform as they overfit and lose generality due to the lack of data. In comparison, our models preserve and slightly increase performance on the domain ($\uparrow 5\%$ mIoU)  while significantly improving performance across other datasets, in some instances even duplicating performance (Pascal-Context $\uparrow 200\%$ mIoU).

\begin{table*}[]
	\centering
	\setlength{\tabcolsep}{2pt}
	\resizebox{\textwidth}{!}{%
		\begin{tabular}{l|l| c c c c c c c c c c c c c c c c c c c | c}
			Method &Venue&\rotatebox[origin=c]{90}{\textit{road} } & \rotatebox[origin=c]{90}{\textit{sidewalk} } & \rotatebox[origin=c]{90}{\textit{building} } & \rotatebox[origin=c]{90}{\textit{wall}} &  \rotatebox[origin=c]{90}{\textit{fence}}& \rotatebox[origin=c]{90}{\textit{pole} } &\rotatebox[origin=c]{90}{\textit{light}}&\rotatebox[origin=c]{90}{\textit{sign} } & \rotatebox[origin=c]{90}{\textit{vegetation}} & \rotatebox[origin=c]{90}{\textit{terrain}*}& \rotatebox[origin=c]{90}{\textit{sky}}& \rotatebox[origin=c]{90}{\textit{pedestrian}}& \rotatebox[origin=c]{90}{\textit{rider}} & \rotatebox[origin=c]{90}{\textit{car}}& \rotatebox[origin=c]{90}{\textit{truck}*} & \rotatebox[origin=c]{90}{\textit{bus}}& \rotatebox[origin=c]{90}{\textit{train}*} & \rotatebox[origin=c]{90}{\textit{motorcycle}} & \rotatebox[origin=c]{90}{\textit{bicycle}}& mean\\\midrule\midrule
			MM\cite{10496388}& T-ITS 24& 88.5& 51.0&87.8&38.6&7.4&52.3&56.3&55.5&87.5&0.0&90.5&73.6&51.0&88.6&0.0&64.4&0.0&54.5&60.4& 53.1\\
			
			DIGA\cite{Shen_2023_CVPR}&CVPR 23&88.5&49.9&90.1&51.4&6.6&55.3&64.8&62.7&88.2&0.0&93.5&78.6&51.8&89.5&0.0&62.2&0.0&61.0&65.8&55.7\\
			MIC\cite{hoyer2023mic}&CVPR 23&84.3&45.6&90.1&48.8&9.2&60.8&66.8&64.4&87.4&0.0&94.4&81.4&58.0&89.7&0.0&65.2&0.0&67.1&64.1&56.8\\
			DCF\cite{chen2024transferring} & ACM 24& 93.4&63.1&89.8&51.1&9.1&61.4&66.9&64.0&88.0&0.0&94.5&80.9&56.6&90.9&0.0&68.5&0.0&63.7&66.6&58.4\\\midrule\midrule
		\rowcolor{Gray}	
            UDA-FROVSS &&96.1&68.7&88.8&52.4&36.5&28.9&47.6&46.5&87.9&0.0&93.0&70.1&38.9&89.1&80.3&80.2&60.2&45.6&58.8&61.5\\\bottomrule
	\end{tabular}}
	\caption{\textbf{State of the art performance comparison of the Synthia-to-Cityscapes UDA setting.} mIoU computed across the 19 Cityscapes categories. Our framework is the only available open vocabulary framework to tackle the Synthia-to-Cityscapes UDA setting. Therefore, we are the only framework capable of segmenting Cityscapes private categories, indicated by ``*''.}
	\label{tab:S2C}
\end{table*}

\paragraph{Comparison with State-Of-The-Art UDA Frameworks}
Table \ref{tab:S2C} compares the state-of-the-art performance for UDA frameworks on the Synthia to Cityscapes setup. Notably, our framework significantly outperforms alternative state-of-the-art frameworks. As our framework is open vocabulary, it handles a major drawback of segmenting all 19 Cityscapes categories despite only seeing 16 during training. Notably, we outperform previous state of the art methods by over 8\%. However, there is still room for improvement. Thin and small categories such as \textit{pole} benefit from \emph{lookup} architectures like \cite{hoyer2022hrda, Shen_2023_CVPR, hoyer2023mic} that process multiple detailed crops of the image to refine fine-grained details of the segmentation. Additionally, we present the current state-of-the-art multimodal framework for UDA (MM \cite{10496388}) whos performance is 15\% worse than ours, highlighting the strong performance of our framework.

\begin{figure}[]
	\centering
	\begin{subfigure}[b]{.5\linewidth}
		\centering
		\includegraphics[clip, trim= 900 350 200 200
		,width=\linewidth]{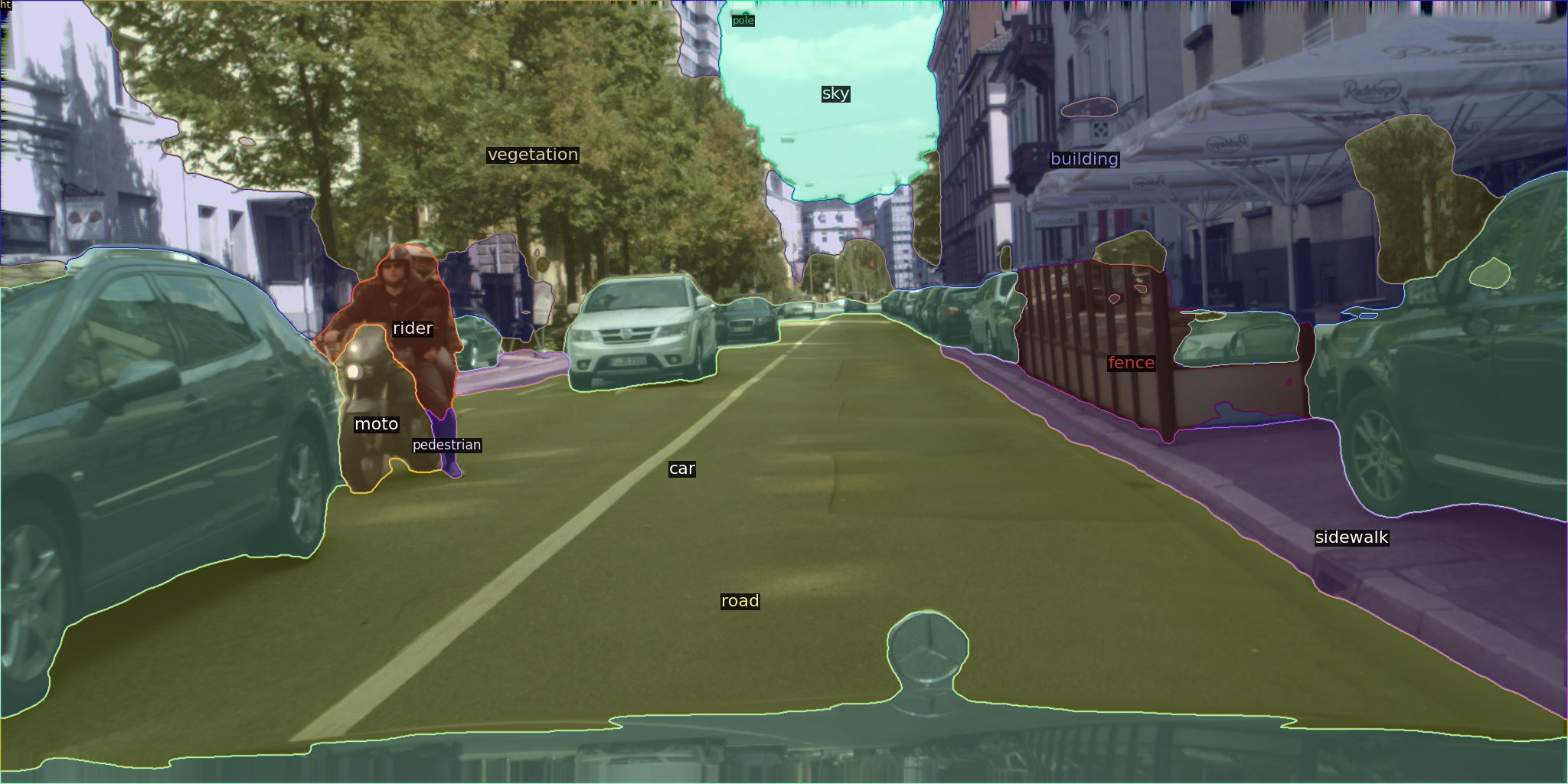}
		\caption{\textit{Fence}.}
		\label{fig:Synthia_fence}
	\end{subfigure}\hfill
	\begin{subfigure}[b]{.5\linewidth}
		\centering
		\includegraphics[clip, trim= 540 400 700 220
		,width=\linewidth]{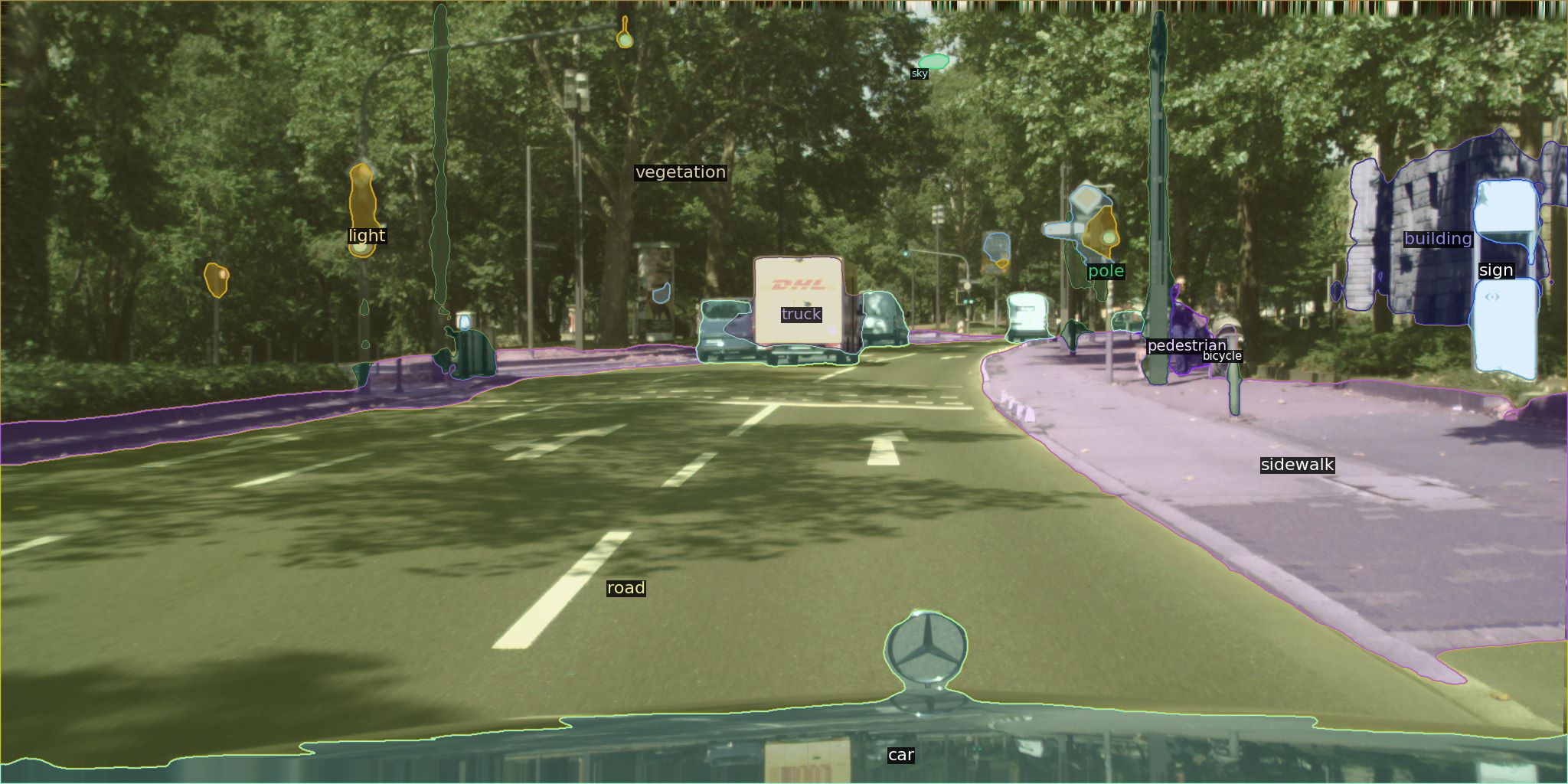}
		\caption{\textit{Truck}.}
		\label{fig:SynthiaTK}
	\end{subfigure}\hfill
	\begin{subfigure}[b]{\linewidth}
		\centering
		\includegraphics[clip, trim= 70 350 800 200
		,width=\linewidth]{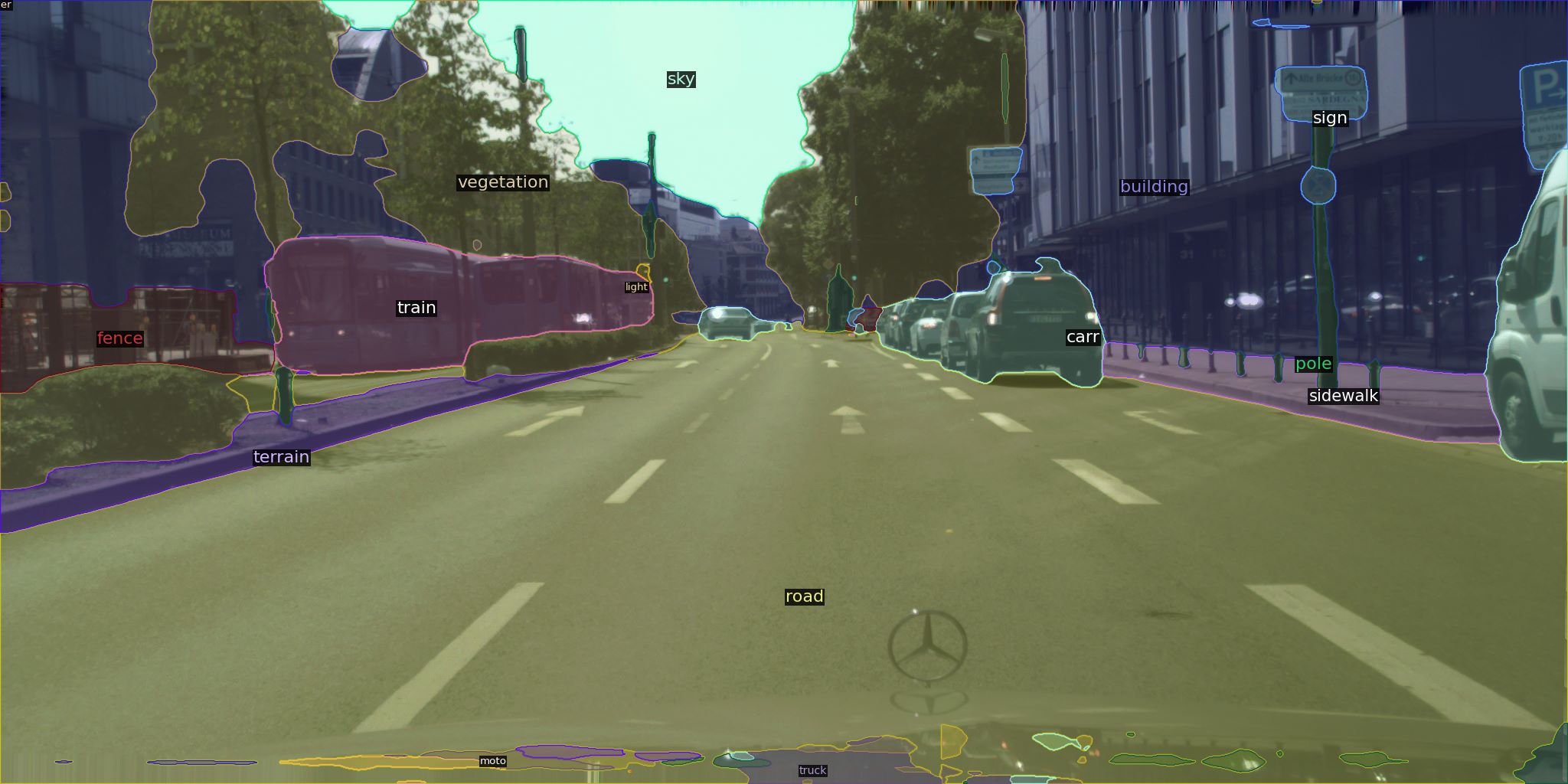}
		\caption{\textit{Train}.}
		\label{fig:SynthiaT}
	\end{subfigure}
	\caption{\textbf{Examples of our UDA-enhanced model in the Synthia-to-Cityscapes setup.} Results on the Cityscapes validation of the model trained with labeled Synthia data and unlabeled Cityscapes images. Segmenting the target private category: \textit{truck} and \textit{train}, and highly unpopulated category: \textit{fence}.}
\end{figure}

Finally, Figures \ref{fig:Synthia_fence}, \ref{fig:SynthiaT} and \ref{fig:SynthiaTK}  qualitatively compare of the segmentation of our UDA model. Notably, it is capable of segmenting categories which state-of-the-art frameworks are not able due to extremely low representation or being absent in the source dataset. 


\paragraph{UDA Improvements Across All Datasets}

\begin{figure}[t]
	\raggedright
	\includegraphics[width=\linewidth]{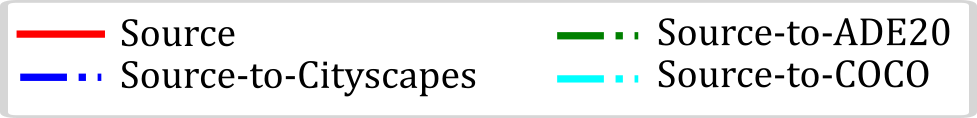}\\
	\begin{subfigure}[b]{0.48\linewidth}
	\includegraphics[width=\linewidth]{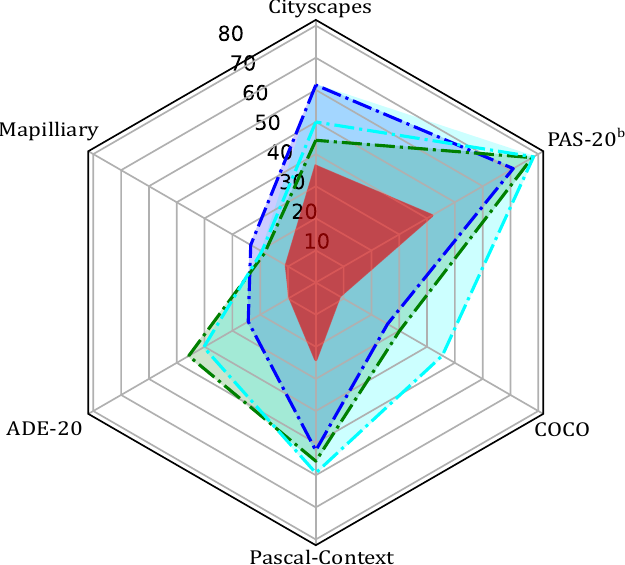}
	\caption{Synthia}
	\end{subfigure}\hfill
	\begin{subfigure}[b]{0.48\linewidth}
	\includegraphics[width=\linewidth]{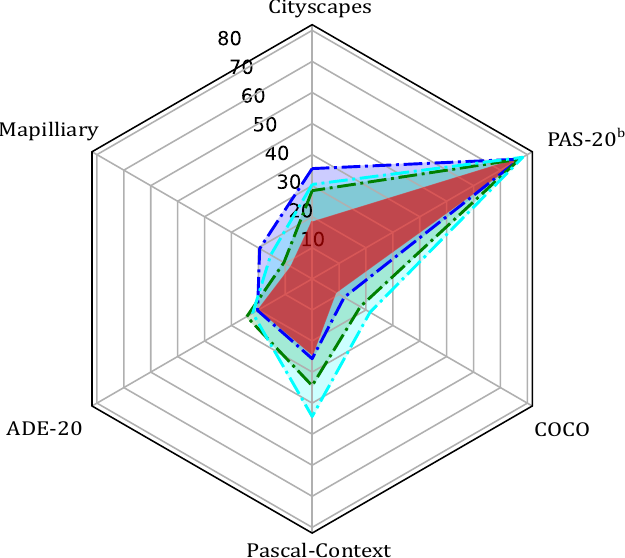}
	\caption{VOC}
	\end{subfigure}
	\caption{\textbf{Performance improvements driven by UDA on small and Synthetic datasets.} In red is depicted the performance of models trained with only source data. Adaptation is performed independently to each analyzed dataset. UDA enhancements correlate with dataset similarity to target domain, as exemplified by significant improvements of PASCAL when performing adaptation with COCO and Mapillary when performing adaptation with Cityscapes.}
	\label{fig:UDA_improvements}
\end{figure}

\begin{figure}[t]
	\centering
	
	\includegraphics[width=.8\linewidth]{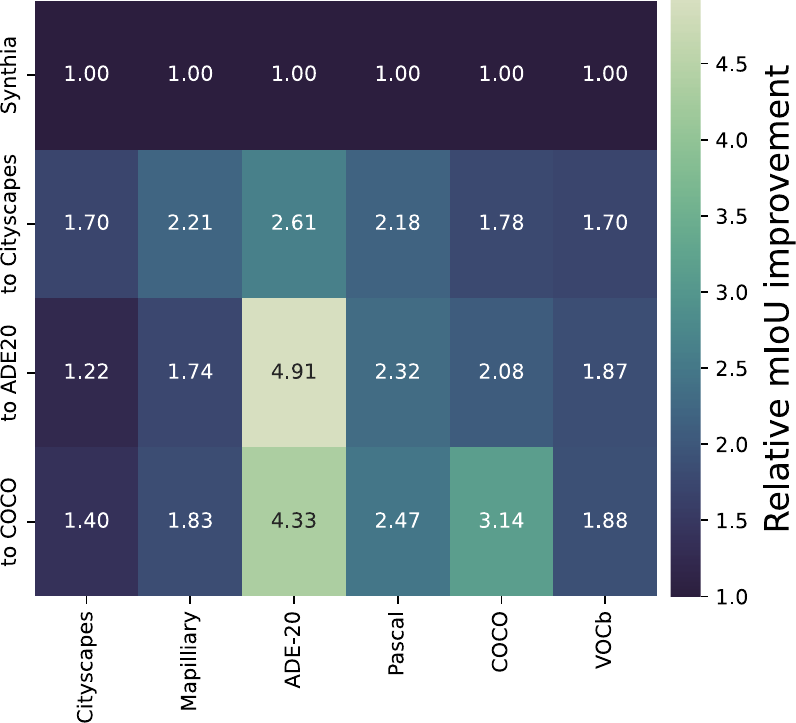}

	\caption{\textbf{Relative performance comparison driven by UDA on a Synthetic dataset.} Relative performance comparison across datasets  of FROVSS  (first row) and UDA-FROVSS both employing  Synthia for training and UDA employing unlabeled images of each dataset. Notably, as Synthia is an urban scenes synthetic dataset the adaptation to a real general purpose dataset (COCO) drives significantly more performance across all datasets.}
	\label{subfig:synthiaUDA}
	\end{figure}

Figure \ref{fig:UDA_improvements} illustrates the performance gains of adapting the student model on the two most challenging datasets (Synthia and PAS-20) to  Cityscapes, ADE-20 and COCO datasets. These results suggest that the domain adaptation techniques not only improve performance on the target set but also improve open vocabulary performance. Additionally, performance improvements on other datasets seem to be highly related to the target dataset employed, see Figure \ref{subfig:synthiaUDA} for a visual relative performance comparison. For example, adapting to Cityscapes provides better performance on Mapillary than adapting to ADE20 or COCO, as both are urban scenes datasets.

\section{Conclusions}
\label{sec:Cc}

In this paper we introduced the first framework for unsupervised domain adaptation open vocabulary semantic segmentation, marking a significant integration of these two research areas. By combining UDA with open vocabulary segmentation, we alleviate the necessity for shared categories between source and target domains, as demonstrated by our improvement over the state-of-the-art UDA frameworks by over 8\% for the Synthia to Cityscapes. Conversely, the open vocabulary approach benefits from UDA's capacity to utilize large volumes of unlabeled data, enabling our models to be successfully trained with less than 2K annotated images. Our approach surpasses previous state-of-the-art for open vocabulary semantic segmentation in all analyzed benchmarks. To achieve these results, we propose a decoder for refined segmentation, a strategic fine-tuning approach to retain CLIP's original weight integrity, and enhanced text embeddings to bolster open vocabulary segmentation. Additionally, we also adapted the teacher-student framework and pseudo-label protocol to effectively train VLMs. For future research, inter-dataset similarity and tuning of the textual encoder emerge as critical factors for further performance enhancements.

\bibliographystyle{CVMbib}
\bibliography{main}



\end{document}